%% file: main.tex
\documentclass[10pt,twocolumn,letterpaper]{article}
\usepackage{etoolbox}

\PassOptionsToPackage{table}{xcolor} 

\usepackage{iccv}              
\usepackage[accsupp]{axessibility}  

\definecolor{iccvblue}{rgb}{0.21,0.49,0.74}
\usepackage[pagebackref,breaklinks,colorlinks,allcolors=iccvblue]{hyperref}
\usepackage{graphicx, multirow}

\usepackage{pifont}  

\def\paperID{13266} 
\def\confName{ICCV}
\def\confYear{2025}

\title{B-VLLM: A Vision Large Language Model with\\ Balanced Spatio-Temporal Tokens}

\author{Zhuqiang Lu\textsuperscript{1},
Zhenfei Yin\textsuperscript{1,2,\dag},
Mengwei He\textsuperscript{1},
Zhihui Wang\textsuperscript{3}, 
Zicheng Liu\textsuperscript{4}, 
Zhiyong Wang\textsuperscript{1},
Kun Hu\textsuperscript{5,\dag}\\
\textsuperscript{1}School of Computer Science, The University of Sydney, NSW 2006, Australia \\
\textsuperscript{2}Department of Engineering Science, University of Oxford, Oxford OX1 2JD, United Kingdom \\
\textsuperscript{3}International School of Information Science and Engineering,\\ Dalian University of Technology, Dalian 116081, China\\
\textsuperscript{4}Advanced Micro Devices, WA 98007, USA\\
\textsuperscript{5}School of Science, Edith Cowan University, WA 6027, Australia\\
\tt\small \{zhuqiang.lu, zhenfei.yin, mengwei.he\}@sydney.edu.au, zhwang@dlut.edu.cn\\
\tt\small zicliu@outlook.com, zhiyong.wang@sydney.edu.au, k.hu@ecu.edu.au
}
\begin{document}
\maketitle
\begingroup
\renewcommand\thefootnote{\fnsymbol{footnote}}
\footnotetext[2]{Corresponding authors.}
\endgroup

\input{sec/0_abstract}  
\input{sec/1_intro}

\input{sec/2_related_work}
\input{sec/3_method}

\input{sec/4_experiments}

\input{sec/5_discussion}
\section*{Acknowledgements}
This work was supported by the Australian Research Council (ARC) Linkage Project \#LP230100294, ARC Discovery Project \#DP210102674 and Edith Cowan University
Science Early Career and New Staff Grant Scheme.

{
    \small
    \bibliographystyle{ieeenat_fullname}
    \bibliography{main}
}

\newpage
\input{sec/6_supp}
\end{document}

%% file: sec/0_abstract.tex
\begin{abstract}
Recently, Vision Large Language Models (VLLMs)  with integrated vision encoders have shown promising performance in vision understanding. They encode visual content into sequences of visual tokens, enabling joint processing of visual and textual data. However, understanding videos, especially long videos, remains a challenge as the rapid growth of visual tokens during video encoding risks exceeding VLLMs' context window length and significantly escalates computational cost. 
To restrict the number of visual tokens, existing VLLMs either: (1) uniformly downsample videos into a fixed number of frames or (2) reducing the number of visual tokens encoded from each frame. 
We argue that the former neglects temporal dynamics in videos, while the latter fails to preserve spatial details within individual frame. 
In this work, we propose Balanced-VLLM (B-VLLM), a novel VLLM framework designed to model task relevant spatio-temporal cues, while restricting the number of visual tokens within the VLLM's context window length. 
Central to our framework is a text-conditioned adaptive frame selection module that dynamically identifies task-relevant frames, which are further de-duplicated with a temporal frame token merging strategy.
The visual tokens of these frames then undergo spatial token sampling and an optional spatial token merging strategy for granular control against the token budget.  
Experiments demonstrate the effectiveness of B-VLLM in balancing the number of frames and visual tokens, moreover, our proposed method introduces 10\% performance gain on MVBench. Code is available at \href{https://github.com/zhuqiangLu/B-VLLM.git}{https://github.com/zhuqiangLu/B-VLLM.git}.
\end{abstract}

%% file: sec/1_intro.tex
\section{Introduction}
\label{sec:intro}
\begin{figure}[t]
  \centering
  \begin{subfigure}{0.49\linewidth}
    \includegraphics[width=\columnwidth]{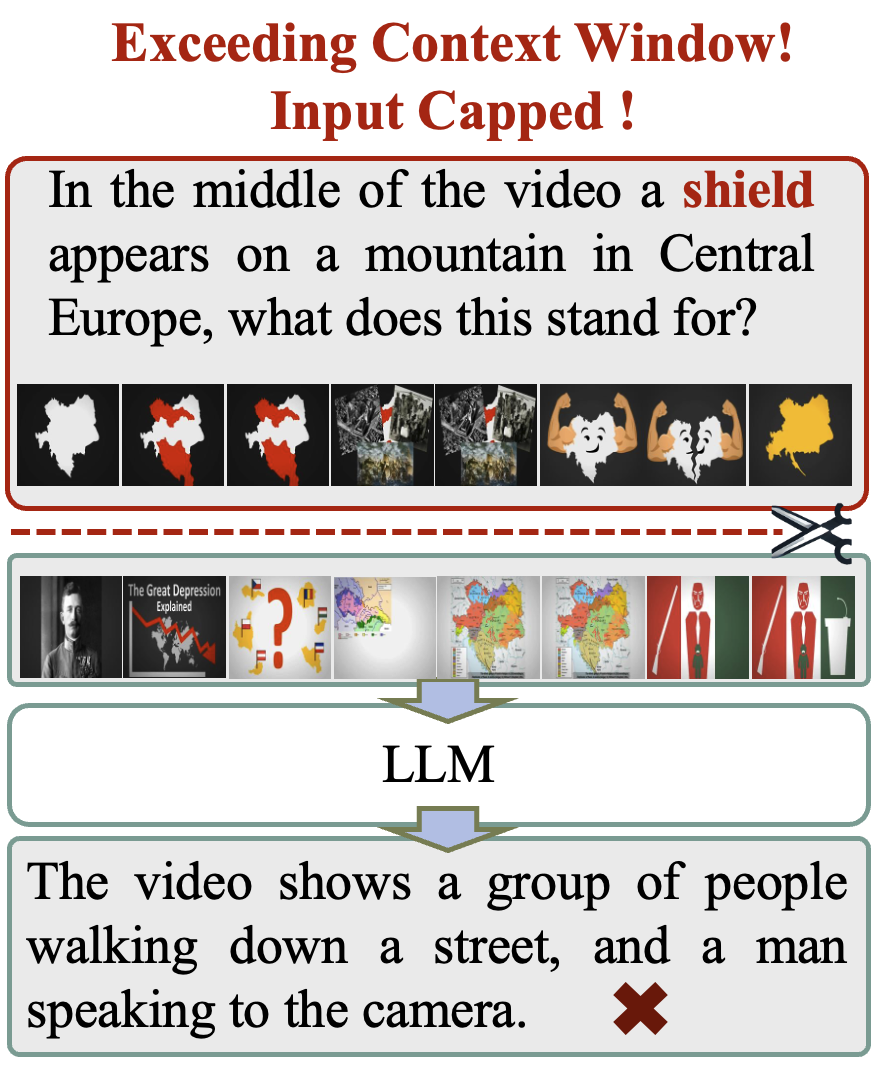}
    \caption{Context window limit.}
    \label{fig:intro-a}
  \end{subfigure}
  \hfill
  \begin{subfigure}{0.49\linewidth}
    \includegraphics[width=\columnwidth]{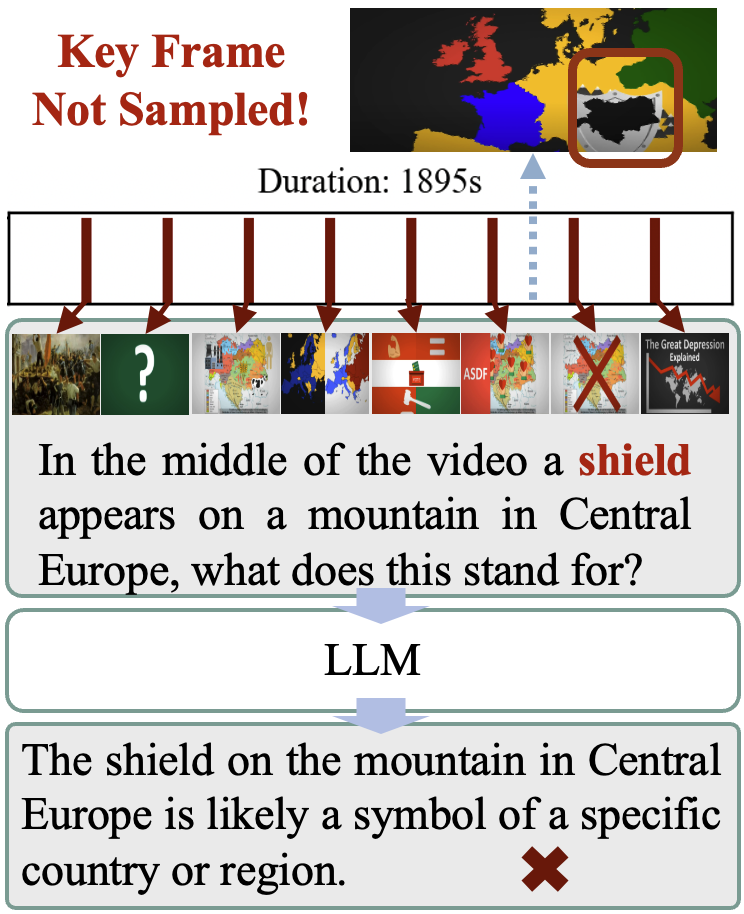}
        \caption{Key-hints missed in sampling.}
    \label{fig:intro-b}
  \end{subfigure}
\vfill
  \begin{subfigure}{0.99\linewidth}
    \includegraphics[width=\columnwidth]{
    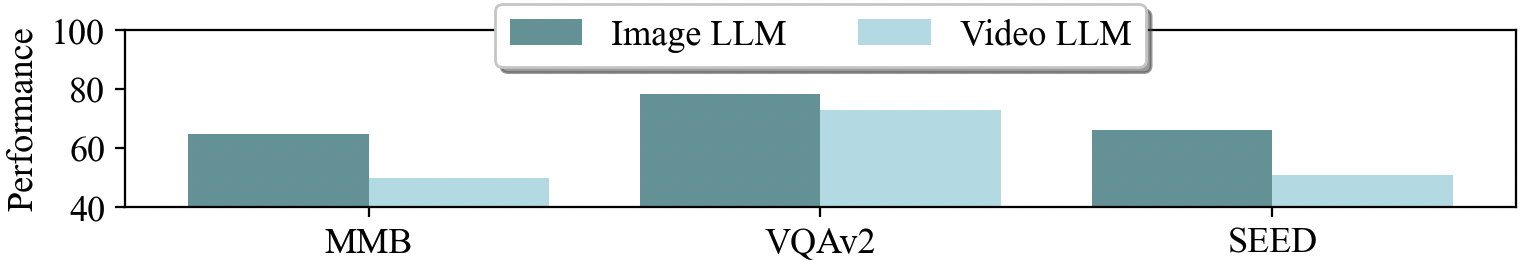}
    \caption{Video LLM performance on image-based spatial-only benchmarks.}
    \label{fig:intro-c}
  \end{subfigure}
  \caption{Illustration of the limitations of existing VLLMs.}
  \label{fig:intro}
\end{figure}
Large Language Models (LLMs) have demonstrated remarkable capability and generalizability in natural language understanding~\cite{qwen, qwen2, llama2, llama3}. Recently, there have been several notable efforts to extend LLMs into Vision Large Language Models (VLLMs) for visual understanding~\cite{flamingo, llava, languagebind}. These efforts primarily aim to adapt LLMs' contextual reasoning capabilities to multimodal settings like video captioning and visual question answering, enabling them to generate accurate textual responses based on visual and textual context provided by users.  
Unlike the extension of vision models to downstream tasks~\cite{hu2018vision, yue2023surgicalpart,lu2024autoregressive, zhao2025clip, chen2023diffusiondet}, bridging the modality gap between textual and visual contexts poses significant challenges~\cite{llava}. Existing VLLMs typically use pretrained Vision Transformers (ViTs)~\citep{vit} to encode images into sequences of visual tokens. These tokens are subsequently integrated with text token sequences as input to the VLLMs~\citep{llava, vila}. This approach compresses spatial contexts from images into 1D sequences and has demonstrated its effectiveness across various computer vision tasks~\citep{mme_benchmark}.
However, these VLLMs are primarily designed for single-image scenarios. As illustrated in Figure~\ref{fig:intro-a}, extending them to video data often compromises performance, as the number of visual tokens exceeds the length limit of the VLLM's context window. 

To address this issue, methods like LLaMA-VID~\citep{llamavid}, MovieChat~\citep{moviechat} and LongVLM~\citep{longvlm} reduce the number of visual tokens to encode each frame. Alternatively, methods such as VideoLLaMA2~\citep{videollama2}, VideoLLaMA~\citep{videollama} and VideoChat2~\citep{videochat2} sample a fixed number of frames uniformly from a video to prevent an overload of visual tokens.
While being effective in controlling the number of visual tokens for video processing, these methods struggle to balance spatio-temporal cues. As depicted in Figure~\ref{fig:intro-b}, VLLMs using uniform frame sampling tend to underperform  on temporally relevant tasks. As shown in Figure~\ref{fig:intro-c}, methods that reduce the number of visual tokens per frame often perform poorly on spatially relevant tasks.

We argue that this is due to a spatio-temporal token imbalance: reducing frame-level visual tokens causes temporal cues to dominate, whereas uniform frame sampling leads to spatial cues to be overshadowed. 
To address these issues, we propose Balanced-VLLM (B-VLLM), a novel VLLM framework to balance spatio-temporal cues in video data conditioned by task-related text prompts. 
B-VLLM processes each frame with both coarse-grained and fine-grained visual tokens. 
A learnable frame selection module leverages coarse-grained tokens to identify the frames most relevant to the video understanding task, guided by the text prompt. 
The fine-grained visual tokens of these selected frames are then processed by a spatial visual token sampling module to retrieve those most task-relevant ones.
For granular control over the final number of generated visual tokens, we introduce an iterative spatial token merging strategy that averages similar visual tokens until the expected number of tokens is reached.
Finally, the resulting visual tokens are mapped into the LLM feature space and integrated with the sequence of textual tokens encoded from the text prompt before being fed into the LLM. 
Our proposed method achieves SOTA performance on various benchmarks.
Notably, our method introduces up to 10\% performance improvement on  MVBench~\cite{videochat2}

The main contributions are summarized as follows:
\begin{itemize}
    \item We propose B-VLLM, a novel Vision LLM framework that dynamically balances spatio-temporal tokens for video understanding, especially for long videos.
    \item We devise an adaptive token selection strategy by leveraging both coarse-grained and fine-grained tokens.
    \item We undertake comprehensive experiments to show that B-VLLM  is effective in different VLLM architectures for video understanding.
    \item We demonstrate that the B-VLLM framework can be generalized to different datasets and VLLM settings.
\end{itemize}

%% file: sec/2_related_work.tex
\section{Related Work}
\label{sec:related_work}
\subsection{Vision Large Language Models}
Vision Large Language Models (VLLMs) extend existing Large Language Models (LLMs) to the domain of visual understanding through visual-instruction tuning~\citep{llava}. Visual content is encoded as sequences of visual tokens using a pretrained CLIP~\citep{clip} vision encoder. These visual tokens are then projected into the LLM hyperspace through a trainable MLP layer~\cite{llava, llamavid, videollama2, videollava}. 
However, in multi-image or video scenarios, the number of vision tokens often exceeds the VLLM's context window limit, leading to performance degrading and high computation costs. 
To address this, Open-Flamingo\cite{flamingo}, MovieChat~\cite{moviechat} and BLIP2 ~\cite{blip2} reduce the number of vision tokens extracted from each frame by using transformer-based token selection networks, such as Q-Former~\citep{blip2} and Perceptive Resampler~\citep{flamingo}. 
However, the number of vision tokens is proportional to the number of input images or frames, creating the risk of an overwhelming number of tokens when processing long videos.
LLaMA-VID~\citep{llamavid} addresses this by encoding each frame into only two visual tokens to enable hour-long video processing; however, this extreme compression rate limits LLaMA-VID's ability to capture finer spatial details in each frame. 
Another common approach to controlling the number of visual tokens is to uniformly sample frames from a video~\cite{llava, videollava, vila, visionllm, videollama2}. 
However, uniformly frame sampling may overlook important frames, especially when keyframes of interests are not evenly distributed throughout the video. 
Compared to existing studies, we propose a vision LLM framework empowered by an adaptive, text-conditioned frame selection mechanism that dynamically selects the most relevant frames for the given vision task.

\subsection{Token Merging Techniques}
Token merging~\cite{tokenmerge} is a technique that combines similar tokens to reduce computational costs with minimum impact on performance in vision tasks. 
Concurrently, SD-TOME~\cite{tome_sd} significantly reduces the computation load of text-to-image generation models through token merging. VID-TOME~\cite{vidtome} demonstrates that token merging is also effective for processing visual tokens in videos. 
In LLMs~\cite{tome_llm}, token merging significantly improves the inference speed by reducing the number of textual tokens. In VLLMs, TOME-LLaVA~\cite{tome_llava} merges visual tokens to lower computational cost. 
More recently, Chat-Uni-VI~\cite{chatunivi} integrates token merging into video LLM pipeline to reduce the number of visual tokens for long videos. LongVLM~\cite{longvlm} divides long videos into segments and applies token merging within each segment's visual tokens. 
Despite these studies integrating token merging into VLLMs, they often lack flexibility in controlling the number of tokens. 
Our study for the first time proposes an iterative token merging strategy to achieve finer control over the number of tokens.

%% file: sec/3_method.tex
\begin{figure*}[t]
  \centering
   \includegraphics[width=\linewidth]{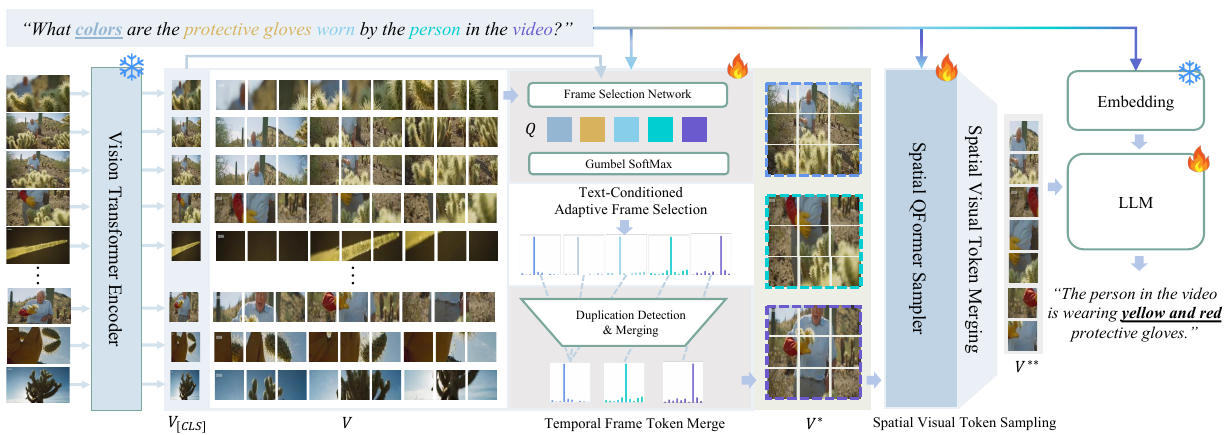}
   \caption{Overview of B-VLLM. The frames of the input video are encoded into preliminary visual tokens $V$ including [CLS] tokens $V_{[CLS]}$ by the Vision Encoder. Subsequently, temporal frame selection and token merging use $V_{[CLS]}$ and the text prompt to identity the visual tokens $V^*$ corresponding to the frames most relevant to the text prompt. Finally, via the spatial sampling and an optional spatial merging strategy, visual tokens $V^{**}$ sampled from $V^*$ are fed into the LLM.}
   \label{fig:overview2}
\end{figure*}

\section{Balanced-VLLM}
\label{method}
In this section, without loss of generality, we assume that the input to Balanced-VLLM (B-VLLM) is a video-text pair. The video consists of $L$ frames, denoted as $I = \{I_1, ..., I_L\}$, where $I_l$ represents the $l$-th frame. The text context input, consisting of $N$ words is denoted as $T = \{T_1, ..., T_N\}$, where $T_n$ is the $n$-th word. 
The goal of B-VLLM is to control the number of visual tokens when processing long videos, thereby minimizing the risk of exceeding the context window limit of the backbone LLM. The overall architecture of the proposed method is described in Sec.~\ref{method_overview}, followed by the \textit{text-conditioned frame selection module} in Sec.~\ref{temporal sampling}, \textit{temporal frame token merging} in Sec.~\ref{temporal merge},  \textit{spatial visual token sampling} and an optional \textit{spatial visual token merging} in Sec.~\ref{spatial sampling}.

\subsection{Architecture Overview}
\label{method_overview}
The model architecture of B-VLLM is depicted in Figure~\ref{fig:overview2}. 
B-VLLM processes the visual tokens of the input video to keep their number below a pre-set threshold $\theta$. 
Initially, each frame $I_l$ is encoded as a sequence of preliminary visual tokens $V_l = \{V_{l,CLS}; V_{l,1}, ..., V_{l,M}\}$ by a vision transformer, where $V_{l,CLS}$ is a [CLS] token that formulates the coarse semantic of the $l$-th frame, $V_{l, j}$ is the $j$-th visual token of $l$-th frame and $M$ is the total number of preliminary frame-level visual tokens except the [CLS] token. In addition, we denote $V = \{V_{1}, ..., V_{L}\}$ to represent the input video with the visual tokens of each frame. 
Unlike previous works~\cite{llava, llamavid, videollama2} that discard [CLS] tokens, we explore using [CLS] tokens in video understanding with LLMs.

First, our text-conditioned frame selection module utilizes the coarse semantics captured by $L$ [CLS] tokens in the video and the given text context $T$ to select the $L^*$ most informative frames. 
The selected frames with their visual tokens can be denoted as: $V^*=\{V^*_1,..,V^*_{L^*}\}\subseteq V$.
To further lower the total number of visual tokens, we utilize a frame-wise token selection module to sample the $Z$ most informative visual tokens from each $V^*_{l^*}\in V^*$, conditioning on $T$.  
An optional spatial token merging module is proposed to combine the selected visual tokens that bear high similarity for a further control in the number of tokens. 
The final selected visual tokens are projected into the LLM feature space and combined with the text context $T$ to form as the input of LLM for downstream tasks, such as \textit{video captioning} and \textit{visual question answering}.

\subsection{Text-Conditioned Adaptive Frame Selection}
\label{temporal sampling}
Using only the frames most relevant to the given text context is essential for reducing the number of visual tokens. However, identifying these frames can be computationally expensive. To address this, we use the [CLS] token $V_{l, [CLS]}$ for each frame instead of including all other preliminary frame-level visual tokens $V_{l, 1}, ...V_{l, N}$. This design choice balances computational efficiency with the effectiveness of locating relevant frames.
We adapt a Q-Former~\cite{blip2} as our frame selection network due to its lightweight design and multimodal reasoning capability. 

In detail, the Q-Former creates $L^*$ queries $Q = \{q_{1}, ..., q_{L^*}\}$ via jointly encoding the [CLS] tokens $V_{[CLS]} = \{V_{1, [CLS]}, ... V_{L, [CLS]}\}$ and the text context $T$. These queries are for locating the $L^*$ most informative frames from $L^*$ perspectives and their corresponding visual tokens. 
Mathematically, we have: 
\begin{equation}
    V^* = S_{\tau} \cdot V = \Phi(Q \cdot V_{[CLS]}^\top,\ \tau) \cdot V,
\end{equation}
where $\Phi$ is Gumbel-Softmax operation and $\tau$ is a temperature parameter for the Gumbel-Max trick. 
Note that $S_{\tau} = \{s_{\tau,l^*,l}\}$ is an $L^* \times L$ matrix, in which $s_{\tau,l^*,l}$ reflects the weight of $l$-th frame for the $l^*$-th selection perspective. As a result, $V^*$ is a linear weighted combination of $V$ regarding the weight $S_\tau$ for $L^*$ perspectives. 

Notably, as $\tau$ approaches 0, the Gumbel-Softamx distribution converges to a categorical distribution, meaning that $S_{\tau=0}$ is a matrix where each row is a one-hot vector. 
Leveraging the property of the Gumbel-Softmax, we construct $\bar{V}$ to enable discretely sampling from $V$. Hence, conceptually, we have $V^*_{l^*}\in V^*$.
In particular, if $s_{\tau=0, l^*, l} = 1$, this indicates that ${V}_{l^*}^* = V_{l}$, meaning $V_{l}$ is selected for the $l^*$-$th$ perspective. 
In this work, we typically set $\tau$ to 0.1 to ensure the a close approximation of one-hot vectors.  Lastly, we restore the temporal order of $V^*$ based on $S_\tau$ as Gumbel-Softmax may break the order of selected frames.
We argue that Softmax is unsuitable in this context as $S_{\tau}$ is calculated based on $V_{[CLS]}$ where each $V_{[CLS], l}$ reflects coarse semantic and may overlook fine-grained spatial details. As a result, Softmax may smooth out rich spatial visual cues especially when aggregating multiple $V_{l}$s. 
\subsubsection{Discussion: Frame Selection with [CLS] Token.}
The role of [CLS] tokens, which carry high-level semantic information of corresponding images~\cite{wang2024cls, cls_closer}, has been widely studied, and our ablation studies in Section~\ref{sec:ablation} validate their effectiveness in frame selection.

\subsection{Temporal Frame Token Merging}
\label{temporal merge}
The text-conditioned adaptive frames selection retrieves the $L^*$ frames with their preliminary visual tokens that are most relevant to the context $T$. However, duplications in the selected frames may occur,
for example, when the number of frames $L$ is less than $L^*$, leading to duplications. 
To address this, we propose a temporal visual token merging mechanism to combine similar or duplicated frames selected. 

Starting with duplication detection, we use $S_{\tau}$ to identify duplicates in the selected frames.
As each perspective (i.e., row-vector) in $S_{\tau}$ approximates a one-hot vector distribution, 
thereby we can measure the row-wise similarity to locate the potential duplications. 
With a cosine similarity measure, two row vectors are marked as duplicates if their similarity exceeds a pre-set threshold $\gamma$. We conduct this procedure in an iterative manner. For 
$V^*_{\alpha} \in V^*$, we identify and update the frames iteratively as a set: 
\begin{equation}
    D_\alpha = \{\beta | \cos(s_\alpha, s_\beta) \geq \gamma, \forall s_\beta \in S \} \cup \{\alpha\},
\end{equation}
where $s_\alpha$ and $s_\beta$ are the $\alpha$-th and the $\beta$-th row vectors in $S$, respectively. 
Next, we update $V^*$ as follows:
\begin{equation}
    V^* \leftarrow  V^* \setminus \{V^*_{\beta}| \beta\in D_\alpha\}.
\end{equation}
The visual tokens of the frames which have indices in $D_\alpha$ are merged as follows:
\begin{equation}
    V^*_{\alpha} = \frac{1}{|D_\alpha|} \sum^{}_{\beta \in D_\alpha}{V^*_{\beta}} .
\end{equation}
Lastly, we append the merged visual tokens back to $V^*$ as:
\begin{equation}
    V^* \leftarrow  V^* \cup \{V^*_{\alpha}\}.
\end{equation}

\subsection{Spatial Visual Token Sampling}
\label{spatial sampling}
Although the maximum cardinality of $V^*$ is limited,  
each $V^*_{i}$ contains $M$ visual tokens except the [CLS] token, resulting in a total of up to $L^*M$ visual tokens. 
Existing studies~\cite{llava, llavanextvideo} typically set $M$ to 256 or 576. When 32 frames are selected, this result in 8,192 or 18,432 visual tokens, respectively, increasing the risk of exceeding the LLM context window limit and imposing a heavy computation load.
To reduce the number of visual tokens per frame, we use a spatial Q-Former to sample $R$ visual tokens that are most relevant to the given context $T$, with $R \ll M$.
In detail, for each $V^*_{l^*}$, the spatial Q-Former generates $R$ learnable spatial visual tokens: $V^{**}_{l^*} = \{V^{**}_{l^*,1}, ...,V^{**}_{l^*, R}\}$ with the inputs $V^*_{l^*}=\{V^*_{l^*,1}, ..., V^*_{l^*, M}\}$ and the text context $T$. Finally, each video contains a total of $MR$ visual tokens. 

Notably, the spatial Q-Former generates a fixed number of tokens, which limits flexibility when finer control over the token count is needed. 
To address this, we propose an optional progressive spatial visual token merging strategy for the scenarios where a maximum visual token count $\theta$ applies.
Simply put, if the size of $V^{**}_{l^*}$ exceeds the maximum token limit, we halve $V^{**}_{l*}$ by merging the most similar tokens through Bipartite Merging~\cite{tokenmerge}. 
This process is repeated until the number of tokens is satisfied.

\subsection{Integration with Backbone LLMs}
As the backbone LLM is a unimodal model that only responds to textual input, we bridge the modality gap between textual and the visual tokens via a learnable MLP model.
Next, the projected visual tokens are concatenated with the textual token sequence before being fed to the backbone LLM. 
Notably, our proposed method is flexible in either integrating with existing VLLM or in being a VLLM on its own. In particular, it is compatible with most of the existing VLLMs architectures, such as LLaMA-VID~\cite{llamavid} and VideoLLaMA2~\cite{videollama2}. Here, we denote our method as B-VLLM when there is no integration with other VLLM. In such case, we utilize Qwen2~\cite{qwen2} as backbone LLM for B-VLLM.

\begin{table*}[ht]
    \small
    
  \centering
  \renewcommand{\arraystretch}{0.9}
  \setlength{\tabcolsep}{5pt}
  \begin{tabular}
    {cc|ccccccc|ccc
  }
     \toprule\
        Method   & \#F &MVB &Perc &VNB &EgoS &VMME$_s$ &VMME$_m$ &VMME$_l$  &MSVD$_{qa}$ &MSRVTT$_{qa}$ &AcNet$_{qa}$ \\
    \midrule
        Video-LLaVA & 8 & 41.0 &  44.3 & - & 38.4  & 45.3 & 38.0 & 36.2 & 70.7/3.9 & \textbf{59.2}/3.5 & 45.3/3.3\\
         Chat-UniVi-V1.5& 8 &- & - & - & - & 45.7 & 40.3 & 35.8 & 69.3/3.7 & 55.0/3.1 & 46.1/3.3 \\
        VideoChat2 & 16 & \underline{}{51.1} &  48.3 & 12.4 & 33.2 & 48.3 & 37.0 & 33.2  & 70.0/3.9 & 54.1/3.3 & 49.1/3.3\\
        Sharegpt4Video & 16 & \textbf{51.2} &  - & - & - & 48.3 & 36.3 & 35.0  & 45.6/- & 43.0/- & 50.8/-\\
        LLaVA-NeXT-V & 32 & 46.5 & \underline{48.8} & - & 43.9 & 36.0 & 32.9 & 28.0 & 67.8/3.5 & -/- & \textbf{53.8}/3.2\\
    \midrule
       LLaMA-VID & 1$^{fps}$ & 39.0 & 41.2 & 5.0 & 31.2  & 34.2 & 34.7 & 27.1 & 69.7/3.7 & 57.7/3.2 & 42.4/3.2\\
      w. Ours & 1$^{fps}$ & 43.5  & 47.0 & 11.6 & 38.2  & 44.7 & 38.8 & 35.2 & 71.5/4.0 & 57.9/3.5 & 43.9/3.4\\
    \rowcolor{lightgray} $\Delta$ & - & +4.5  & +5.8 & +6.6 & +7.0  & +10.5 & +4.1 & +8.1 & +1.7/+0.3 & +0.2/+0.3 & +1.5/+0.2\\
       Video-LLaMA2 & 8 & 45.5  & 45.6 & 10.8 & 42.2 & \underline{48.9} & 42.7 & 37.7  & 70.2/3.9 & 54.8/3.3 & 47.6/3.4\\
      w. Ours & 1$^{fps}$ & 46.5  & 48.0 & 12.6 & 44.3 & 47.2&  \underline{44.4}& \underline{41.5}  & \underline{72.1}/4.0 & 58.1/3.5 & 48.5/3.4\\
      \rowcolor{lightgray}$\Delta$ & - & +1.0  & +2.4 & +1.8 & +1.9  & -1.7 & +1.7 & +3.8 & +1.9/+0.1 & +3.3/0.2 & +0.9/+0.0\\
      \midrule
     B-VLLM & 1$^{fps}$ & 50.8  & \textbf{52.1} & \textbf{22.6} & \textbf{51.9} & \textbf{60.8} & \textbf{51.8} & \textbf{47.9}  & 
     \textbf{72.3}/4.0 & \textbf{59.2}/3.5 & 48.6/3.7\\
      
    \bottomrule

  \end{tabular}
  \caption{Quantitative comparisons on video benchmarks in terms of accuracy $\uparrow$.
  }
  \label{tab:video_mcqa}
\end{table*}

\section{Implementation} 
\textbf{Baselines}. We evaluate against two baselines: LLaMA-VID~\cite{llamavid}, which uses high spatial compression for video understanding, and VideoLLaMA2~\cite{videollama2}, which uniformly samples 8 frames per video.
\noindent\textbf{Datasets \& Training}.  B-VLLM is trained solely on LLaMA-VID-Dataset~\cite{llamavid} and Valley~\cite{valley} for fair comparison. See supplementary material for details.

%% file: sec/4_experiments.tex
\begin{table*}[]
    \small
    
    \setlength{\tabcolsep}{9pt}
    \renewcommand{\arraystretch}{0.9}
  \centering
  \begin{tabular}{cc|ccccccccc}
     \toprule\
        Method   & \# Token. & MMB &  POPE & VizWiz & MME & VQAv2 & SciQA & GQA & TQA& SEED \\
       
    \midrule
    InstructBLIP & 32 & 36 &  - &  34.5 & - & - & 60.5 & 49.2  & 50.1  & 58.8\\
    Qwen-VL & 256 & 38.7 &  - &  35.2 & - & \textbf{78.8} & 67.1 & 59.3  & \textbf{63.8}  & 56.3\\
    Qwen-VL-Chat & 256 & 60.6 &  - &  38.9 & 1488 & 78.2 & 68.2 & 57.3  & \underline{61.5}  & 58.2\\
    LLaVA-1.5 & 576 & 64.7 &  \textbf{85.6} &  \textbf{50.0} & \textbf{1510} & \underline{78.5} & 66.8 & \textbf{62.0}  & 58.2 & 66.1\\
    \midrule
    LLaMA-VID & 2 & 49.8 &  75.3 &  37 & 1187 & 62.7 & 65.3 & 48.8  & 42.1 & 50.8\\
    w. Ours  & 32 & 59.3 &  83.8 &  36.9 & 1285 & 73.0 & 66.9 & 53.0  & 46.6 & 61.0 \\
    \rowcolor{lightgray} $\Delta$  & - & +9.5 &  +8.5 &  -0.1 & +87 & +10.3 & +1.3 & +4.2  & +4.5 & +10.2 \\
    VideoLLaMA2  & 72 & 64.6 &  84.3 &  46.7 & 1455 & 74.5 & \underline{72.2} & 56.9  & 54.6 & 63.9 \\
    w.Ours  & 32 & \underline{67.3} &  \underline{85.3} &  \underline{47.3} & 1501 & 74.5 & 72.0 &  59.7 & 52.8 & \underline{66.7} \\
    \rowcolor{lightgray}  $\Delta$  & - & +2.7 &  +1.0 &  +0.6 & +46 & 0.0 & -0.2 & +2.8  & +1.8 & +2.8 \\
    \midrule
     B-VLLM  & 32 & \textbf{71.9} &  84.5 &  45.1 & \underline{1455} & 76.6 & 
     \textbf{74.5} &  58.4 & 54.8 & \textbf{67.1} \\
    \bottomrule

  \end{tabular}
  \caption{Quantitative comparisons on image benchmarks in terms of accuracy $\uparrow$.
  }
  \label{tab:image_bench}
\end{table*}
\section{Experimental Result \& Discussion}
\label{exp}

\subsection{Video Benchmark Quantitative Evaluation}
We evaluate the proposed B-VLLM on various video benchmarks in a zero-shot manner. The video duration in the these benchmarks spans from 10 second to 1 hour, where the details are summarized in the supplementary materials. 
For open-end answering, we have MSVD-QA~\cite{msvd_and_msrvtt}, MSRVTT-QA~\cite{msvd_and_msrvtt} and ActivityNet-QA~\cite{activitynet}, which aim to evaluate the capacity of video captioning.
For multi-choice question answering (MCQA), we have 
MVBench~\cite{videochat2} and VideoMME~\cite{videomme}, which are multi-purpose video understanding benchmarks; EgoSchema~\cite{egoschema} focuses on evaluating the temporal understanding capability;  Perception~\cite{perceptiontest} aims to examine reasoning and perception skills of VLLMs; and
VNBench~\cite{vnbench} evaluates a VLLM using the needle in a haystack manner: asking questions about one or more irrelevant images that are randomly inserted into a video. 

Quantitative results for open-end question answering and MCQA benchmarks are reported in Table~\ref{tab:video_mcqa}.
Overall, B-VLLM achieves SOTA performance on MSVD-QA, Perception, VNBench, Egoschema, and VideoMME. Specifically, when integrating with existing VLLMs, it significantly boosts performance to the two baselines across most video benchmarks. Notably, our method achieves substantial improvements in long-video scenarios. 
For example, VideoMME-Long evaluates VLLMs performance in extremely long-video settings, with an average video duration of 2466.7 seconds. 
Our proposed method introduced an 10.5\% performance gain for LLaMA-VID~\cite{llamavid} and 3.8\% for VideoLLaMA2~\cite{videollama2}. 
Context window length is one of the key factors limiting existing VLMs to handle long videos. For example,
 LLaMA-VID encodes each frame with 2 tokens, it struggles with extremely long video due to its context window limit and potential key information loss. 
Our adaptive frame selection module addresses this limitation by selecting a fixed number of frames most relevant to the given context, ensuring the number of visual tokens remains within the context window of LLaMA-VID. As a result, LLaMA-VID consistently performs well in medium and long video settings. 
We also observe a noticeable performance drop when VideoLLaMA2 handles medium and long videos, which we attribute to its uniform frame sampling strategy that may overlook the frames relevant to the context. The integration of our method mitigates this issue by filtering out irrelevant frames, leading to 1.7\% and 3.8\% performance gains for VideoLLaMA2.

Additionally, the proposed method remains effective in the short-video settings. For example, LLaMA-VID achieves a 4.5\% and 5.8\% performance gain in MVBench and Perception through the integration of our method. Similarly, in open-end benchmarks, performance improvements of 1.7\%, 0.2\% and 1.5\% are observed in MSVD-QA, MSRVTT-QA and ActivityNet-QA, respectively.
This substantial improvement can be attributed to the enhanced spatial information from the increased number of spatial visual tokens compared to LLaMA-VID. 
Similar performance improvements are also observed in Video-LLaMA2, with gains of 1.0\%, 2.4\%, 1.8\%, 2.1\%, 1.9\%, 3.3\% and 0.9\%  on MVBench, Perception, VNBench, Egoschema, MSVD-QA, MSRVTT-QA and ActivtiyNet-QA, respectively. 

An interesting observation is that VLLMs employing uniform sampling achieve better performance in short-video scenarios, particularly when the number of sampled frames closely matches the video length in seconds. For example, VideoChat2~\cite{videochat2} ranks highest on MVBench (16 frames), and LLaVA-Next-Video~\cite{llavanextvideo} ranks second on Perception (23 frames). Although these methods could theoretically improve performance in longer videos by sampling more frames, they quickly exceed the context-window limitations, increasing computational costs and reducing performance. In contrast, our method maintains consistent performance across short and long videos due to effective frame selection and spatio-temporal merging techniques. Refer to the supplementary material for more experimental result.

\subsection{Image Benchmark Quantitative Evaluation}

Following~\cite{llava}, we select MMBench~\cite{mmbench}, POPE~\cite{pope}, VizWiz~\cite{vizwiz}, MME~\cite{mme_benchmark}, VQAv2~\cite{vqa}, ScienceQA-Image~\cite{scienceqa}, GQA~\cite{gqa}, TextQA~\cite{textvqa} and Seed-Image~\cite{seedbench} to evaluate the generalizability of B-VLLM in the image setting. 
Evaluation results are reported in Table~\ref{tab:image_bench}. 

Our proposed method yields comparable, if not superior, performance, despite not being specifically designed for the image-only setting and using \textbf{significantly fewer visual tokens per image than existing image-based VLLMs}. 
In particular, B-VLLM outperforms classic image-only VLLM, LLaVA-1.5~\cite{llava} on MMBench~\cite{mmbench}, ScienceQA-Image~\cite{scienceqa}, and SEED-Image~\cite{seedbench} by 7.1\%, 7.7\%, and 1.0\%, respectively. 
Notably, comparable results are observed on POPE~\cite{pope} and MME~\cite{mme}, despite LLaVA-1.5~\cite{llava} consuming 18 times more visual tokens than ours. 
These benchmarks primarily focus on reasoning and general image understanding, where fine-grained visual details are less critical and typically require fewer visual tokens. 
However, as noted by LLaMA-VID~\cite{llamavid}, more visual tokens generally lead to better performance on certain image-only benchmarks. 
The limitation in the number of visual tokens compromises our method's performance on VizWiz~\cite{vizwiz}, VQAv2~\cite{vqa} and TextVQA~\cite{textvqa}, where these benchmarks examine spatial reasoning and OCR. For example, TextVQA~\cite{textvqa} focuses on detecting text within image, while GQA~\cite{gqa} emphasizes complex scene understanding.

In terms of performance gain, integrating our proposed method results in significant improvements for LLaMA-VID~\cite{llamavid} and VideoLLaMA2~\cite{videollama2} on image benchmarks. 
For LLaMA-VID, 
we observe gains of 9.5\% and 10.2\% on MMB and VQAV2; 
and for VideoLLaMA2, 
non-trivial gains can be observed when incorporating with our method. 
We believe these gains stem from the effectiveness of the visual token sampling module, as it dynamically selects the most informative tokens, reducing the overall token count while maintaining performance.

\begin{table}[hb]
\small  
 \setlength{\tabcolsep}{7pt}
    
  \centering
  
  \begin{tabular}{ccc|cccc}
     \toprule\
        F.S. & T.M. & S.S.  & MVB & VMME  & MMB & POPE \\

    \midrule
      &  &  &  39.0 & 33.6 &49.8 &75.3 \\
      \checkmark &  &  & 39.4 & 36.5  & 58.0 & 81.1 \\
       \checkmark & \checkmark &  &  42.1 & 38.1  & 54.6 & 67.3\\
        \checkmark & \checkmark & \checkmark &  \textbf{43.5} & \textbf{39.6}   & \textbf{59.3} & \textbf{83.8}\\
    
    \bottomrule
  \end{tabular}
  \caption{Ablation study on adaptive frame selection (F.S), temporal frame token merging (T.M.), and spatial visual token sampling (S.S). Experiments are done on LLaMA-VID.
  }
  \label{tab:ablation}
\end{table}

\begin{figure}[t]
  
  \centering
  \begin{subfigure}{\linewidth}
    \includegraphics[width=\columnwidth]{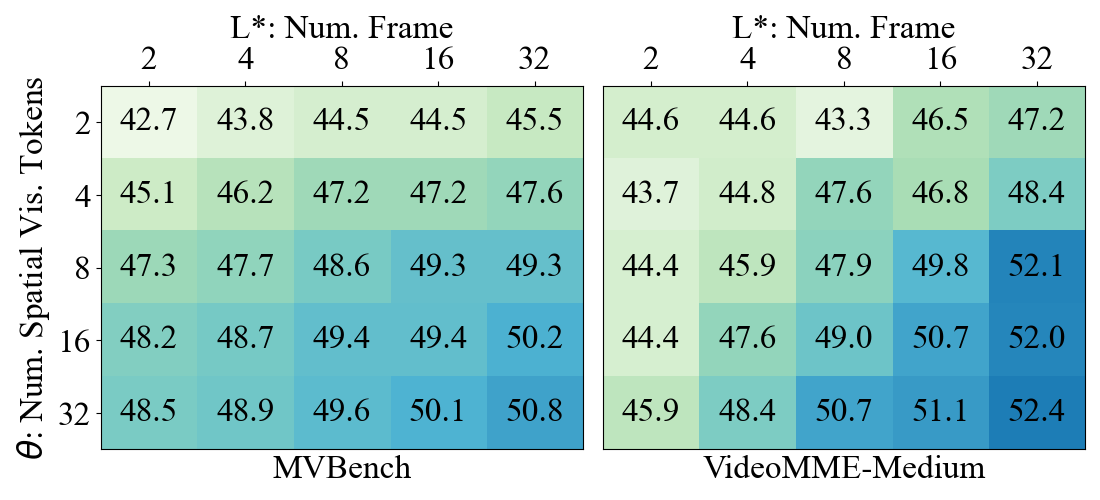}
    \caption{Impacts on the number of frames selected $L*$ and spatial
vis. tokens $R$}
    \label{fig:ab-a}
  \end{subfigure}
  \hfill
  \begin{subfigure}{\linewidth}
    \includegraphics[width=\columnwidth]{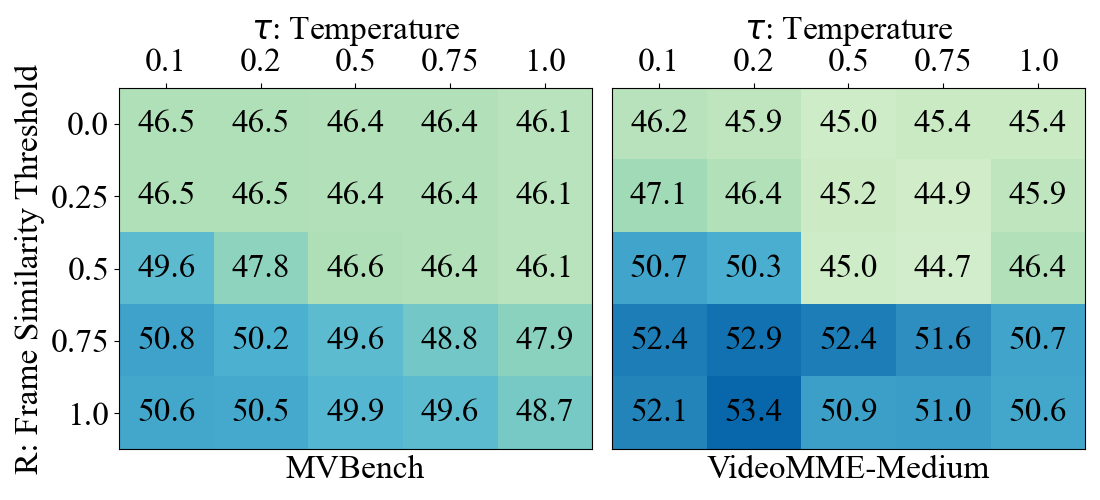}
    \caption{Impacts on the temperature $\tau$ and  frame similarity threshold $\gamma$}
    \label{fig:ab-b}
  \end{subfigure}
    \vfill
    \vfill
  \begin{subfigure}{\linewidth}
    \includegraphics[width=1\columnwidth]{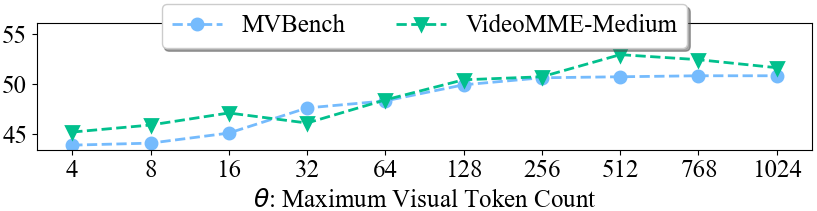}
    \caption{Impacts on maximum number of token $\theta$}
    \label{fig:ab-c}
  \end{subfigure}
  \caption{Impacts on the hyperparameters.}
  \label{fig:ab}
\end{figure}
  

\begin{figure*}[ht]
  \centering

   \includegraphics[width=\linewidth]{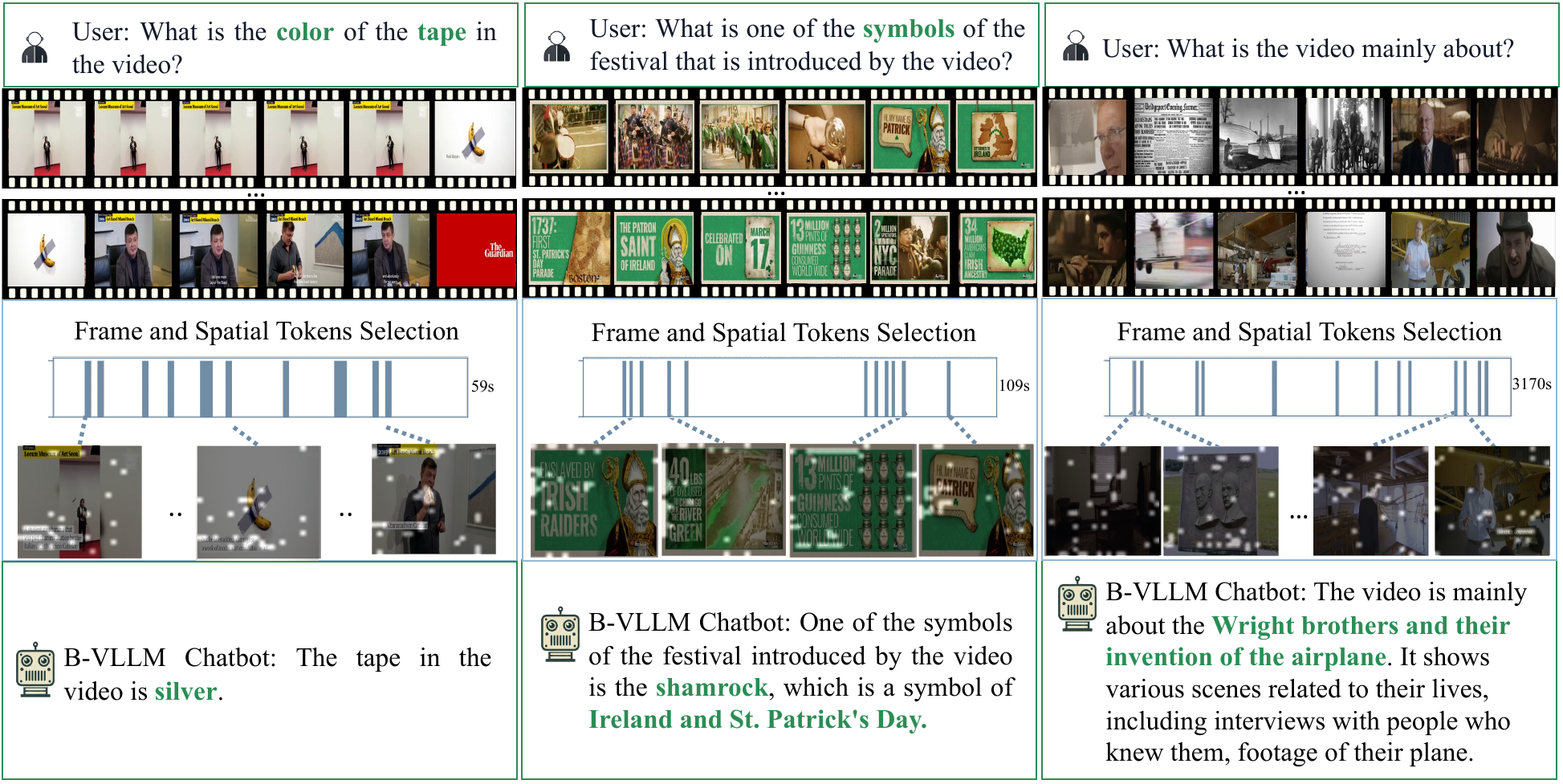}

   \caption{Qualitative examples for two short-medium videos and one long video.}
   \label{fig:vis_1}
\end{figure*}
\subsection{Ablation Study}

\textbf{Effectiveness of the proposed modules} in terms of their performance is reported in Table~\ref{tab:ablation}. 
In this experiment, LLaMA-VID~\cite{llamavid} is selected as the base model. MVBench~\cite{videochat2} and VideoMME~\cite{videomme} are used to evaluate video understanding performance, while MMBench~\cite{mmbench} and POPE~\cite{pope} are used to assess image understanding performance. 
Starting with the baseline, without the integration of our proposed modules, LLaMA-VID~\cite{llamavid} displays moderate performance on the short-video benchmark MVBench~\cite{videochat2}. However, compromised performance can be observed on the medium-to-long video benchmark Video-MME~\cite{videomme} due to the limitation of the context window length. 
In addition, LLaMA-VID~\cite{llamavid} typically performs worst on image benchmarks, due to its poor spatial perception caused by a deficit of spatial visual tokens. With the introduction of the text-conditioned adaptive frame selection (F.S.) module, the baseline model no longer suffers from its own token selection mechanism, resulting in 2.9\% performance boost on the medium-to-long video benchmark VideoMME~\cite{videomme}. 
With the temporal frame token merging (T.M.) mechanism, the model mitigates temporal redundancy in duplicated frame selection, resulting in noticeable performance gains on both short and medium-to-long video benchmarks. 
Similarly, the integration of spatial visual token sampling (S.S.) enables dynamic adjustment on the number of spatial visual tokens to balance the temporal and spatial cues, resulting in performance gains 1.4\% and 1.5\% on MVBench and VideoMME, respectively.   More ablation experiments can be found in supplementary material.

\begin{table}[ht]

    \centering
    \small
    \begin{tabular}{cc|ccccc}
        \hline
       \textbf{Ach./Token.} & \textbf{Time} &\textbf{MVB} & \textbf{EgoS} & \textbf{Perp.} & \textbf{VMME} \\ 
       \hline
       Random & - & 27.7 & 20.0 & 33.0 & 25.0\\
       \hline
       Q-Former & 25 hrs. & 46.5 & 44.3 & 48.0 & 44.5\\
       Resampler& 23 hrs. & 44.3 & 45.5 & 46.8 & 44.0\\
       \hline
     Mean Pooling & 19 hrs.  & 41.2  & 40.0 & 41.1 & 37.6\\
       CLS Token & 10 hrs. & 34.5 & 34.1 & 37.8 & 36.8\\
       \bottomrule
    \end{tabular}
    \caption{Q-Former v.s Resampler in Frame Selection and Mean Pooling v.s [CLS] tokens in VLLMs.
  }
    \label{ablation:qform_vs_resampler_vs_cls_vs_patch}
\end{table}
\noindent\textbf{Impacts on the number of frame selected $L*$ , spatial visual tokens $R$  and  maximum visual token count $\theta$} are illustrated in Figure~\ref{fig:ab-a} and ~\ref{fig:ab-c}. B-VLLM is selected as base model in the upcoming experiments.
First, we examine the impact of two variables: the number of selected frames $L*$ and the number spatial visual tokens $R$ per frame. More specifically, we investigate how different combinations of these variables affect the performance of a trained model.
Interestingly, as shown in Figure~\ref{fig:ab-a}, we observed that, compared to the maximum setting with 32 selected frames and 32 spatial visual tokens per frame, it exhibits only a marginal performance drop when the number of selected frames is reduced while maintaining a sufficient number of visual tokens. 
For example, when the number of spatial visual token $R$ is set to 32, reducing the number of selected frames $L*$ from 32 to 16 results in a performance drop of 0.9\% and 1.3\% on MVBench and VideoMME-Medium, respectively. 
In a more extreme setting, reducing the number of frame selected $L*$ from 16  to 8 results in a performance drop of 0.5\% and 0.4\% performance drop on MVBench and VideoMME-Medium, respectively. 
We argue that this is because 8 to 32 key frames are sufficient for most video understanding tasks. 
When using 32 frames, dropping the number of visual tokens $R$ from 32 to 16 results in performance drops of 0.6\% on MVBench and 0.4\% on VideoMME-Medium. This indicates that current video understanding tasks tend to rely more on temporal dynamics rather than on detailed spatial information. 
Secondly, we examine how the maximum token count $\theta$ affects the trained model while keeping both the number of selected frames $L*$ and the number of spatial visual tokens $R$ fixed at 32. As shown in Figure~\ref{fig:ab-c}, increasing the token count generally improves performance, but the gains for B-VLLM plateau at 512 tokens. This indicates  B-VLLM can deliver strong performance without a high computational budget.
 
\noindent\textbf{Impacts on temperature $\tau$, similarity threshold $\gamma$ } are presented in Figure~\ref{fig:ab-b}. In these experiments, the number of selected frames $L*$ and spatial visual tokens $R$ are both set to 32, and the maximum token count $\theta$ is fixed at 768. Our results indicate that the optimal performance is achieved with a lower $\tau$ value in combination with a higher $\gamma$ value. Where $\tau$ governs the discreteness of Gumbel-Softmax during frame selection and higher $\gamma$ values prevents dissimilar frames from being merged.
For example, when $\tau$ is set to 0.1, reducing $\gamma$ from 0.75 to 0.5 results in 1.2\% and 1.7\% performance drop on MVBench and VideoMME-Medium, respectively. Similarly, when $\gamma$ is fixed to 1.0, lowering $\tau$ from 0.1 to 0.5 results in 0.7\% and 1.2\% performance drops. 
We find that a high $\gamma$ prevents merging dissimilar frames, while a low $\tau$ mitigates the oversmoothing of frame features caused by the Gumbel-Softmax.

\noindent\textbf{Q-Former v.s. Resampler in Frame Selection}. 
Besides Q-Former, we also verified with other commonly used frame selection module - Resampler, where the results are reported in Table \ref{ablation:qform_vs_resampler_vs_cls_vs_patch}. Resampler is originally designed for visual token abstraction, for fair comparison, we provide it with both visual and text tokens, setting the same number of query tokens as in Q-Former.  Notably, Q-Former outperforms Resampler on most selected benchmarks. We argue that the Resampler's reliance on self-attention mechanism limits its performance in processing cross-modality tokens. 

\noindent\textbf{Efficacy of [CLS] token.} To validate the efficacy of [CLS] tokens in frame selection, we compare its performance against using visual tokens with mean pooling as frame features. For simplicity, we conduct preliminary experiment on VideoLLaMA2~\cite{videollama2} where the modality projector is replaced with a single MLP layer. It is worth mentioning that, none of the proposed module is present in this preliminary experiment. All experiments are trained using the same settings, with 8 frames selected from each video. 
As shown in Table~\ref{ablation:qform_vs_resampler_vs_cls_vs_patch}, [CLS] tokens enables the model to consistently outperform random guessing while being more training-efficient than mean pooling. This efficiency is preferable and the effectiveness is sufficient for frame selection, as the goal is to filter out most irrelevant frames in a coarse stage. Refer to the supplementary material for more details.

\label{sec:ablation}
\subsection{Discussion}
\noindent\textbf{Visualization.} As shown in Figure~\ref{fig:vis_1},  we visualize frame selection and spatial tokens for two short-medium videos and one long video. In the first case, the frame selection module locates the frames with the banana taped to the wall. Then based on the question, the visual tokens near the banana and tape are sampled and used by LLM, Lastly, the LLM backbone answers the question correctly based on the selected visual tokens. 
In the second video, the selected frames are from both ends, demonstrating question-based selection rather than random or uniform sampling. In contrast, for long videos with overarching questions, the frames selected are evenly distributed. \noindent\textbf{Limitation.} We recognize that using [CLS] tokens for frame selection has its limitations since pooled visual tokens are more informative, as an alternative strategy for exploring more accurate frame selection.  See the supplementary material for further discussion.

%% file: sec/5_discussion.tex
\section{Conclusion}
We present a novel vision LLM framework - B-VLLM, designed to address the issue of visual token overload in video understanding with LLMs.  Our approach controls the number of visual tokens by adaptively selecting the most relevant frames and applying dynamic spatial token sampling and merging.
Comprehensive experiments confirms the effectiveness of B-VLLM. 
We also highlight the limitations of our method to inform future directions.


%% file: sec/6_supp.tex
\setcounter{page}{1}
\maketitlesupplementary

\begin{table}
    \caption{Statistics of training datasets. 
  }
   \setlength{\tabcolsep}{3pt}
  \centering
  \begin{tabular}{c|c|c|c}
     \toprule\
     
        Stage & Modality & Sample & Source  \\
    \midrule
    \midrule
    \multicolumn{4}{c}{LLaMA-VID~\cite{llamavid}} \\
    
    \midrule
       \multirow{2}{*}{ PT } & Image & 558K  & ~\citep{cc3m, llava} \\
     \  &  Video  & 232K &  \citep{webvid, llamavid} \\
     \midrule
     \  \multirow{3}{*}{ FT }   & Image  & 625K &  ~\cite{vqa, gqa, referitgame, vgenome, uod, orcvqa, aokvqa, textcaps} \\ 
     \  &  Video  & 98K &  ~\cite{activitynet} \\
     \  &  Text  & 40K &  ~\cite{sharegpt} \\
     \midrule
     \midrule
     \multicolumn{4}{c}{VideoLLaMA2~\cite{videollama2} - Valley~\cite{valley}} \\
     \midrule
       \multirow{2}{*}{ PT } & Image & 558K  & ~\citep{cc3m, llava} \\
     \  &  Video  & 702K &  \citep{valley} \\
     \midrule
     \  \multirow{3}{*}{ FT }   & Image  & 665K &  ~\cite{llava} \\ 
     \  &  Video  & 100k &  ~\cite{videochatgpt, valley} \\

    \bottomrule
  \end{tabular}
  \label{tab:dataset}
\end{table}

\begin{table}
    \caption{Statistics of benchmark datasets. Type indicates the QA question type: \textbf{Open} represents open-ended answering and \textbf{MCQA} denotes multi-choice question answering.
  }
  \centering
  \begin{tabular}{c|c|c|c}
     \toprule\
        Benchmark & Avg. Len (s)  & \# QA Pair & Type  \\
    
    \midrule
        MSVD-QA & 9.8 & 10K & Open \\
        MSRVTT-QA & 15.2 & 50.5K & Open \\
       MVBench & 16 & 4K & MCQA \\
       Perception-Test & 23 & 2.7K & MCQA \\
     Next-QA & 44 & 4000 & MCQA \\
     VNBench & 54 & 5.6K & MCQA \\
     VideoMME-S & 80.7 & 0.9K & MCQA \\
    ActivityNet-QA & 111 & 8K & Open \\
     EgoSchema & 180 & 5K & MCQA \\
    VideoMME-M & 515.9 & 0.9K & MCQA \\
    VideoMME-L & 2466.7 & 2.7K & MCQA \\
     
    \bottomrule
  \end{tabular}
  \label{tab:video benchmark}
\end{table}
\section{Additional Implementation Details}
\subsection{Training Details}
We adopt a two-stage training strategy~\citep{llava, llamavid, videollama2}, dividing training into pretraining for modality alignment and fine-tuning for instruction tuning. During pretraining, the learning rate is set to 1e-3 with a linear learning rate schedule, and the total batch size is set to 256. 
Only the frame selection module and the MLP projection layer, which maps the visual token features to the LLM features, are updated.
In the fine-tuning stage, the entire framework is unfrozen except for the preliminary visual encoder. Additionally, LoRA~\cite{lora} is enabled with the rank set to 128 and $\alpha$ set to 256. The total batch size was set to 128, and the learning rate is 1e-4 with cosine scheduling. Supervision is provided by minimizing cross-entropy for masked text tokens, and the optimizer for both stages is AdamW~\cite{adamw}, with DeepSpeed ZeRO2~\cite{deepspeed} enabled for memory efficiency. Training is conducted with 16 Nvidia L40S GPUs.
Following previous works~\cite{videochat2, videochatgpt, videollama2}, videos are downsampled to 1 \textit{fps} in both training and evaluation.

\label{sec:supp_dataset}
\subsection{Dataset for Training}
The following provides details on the training datasets utilized in LLaMA-VID-Dataset~\cite{llamavid} and VideoLLaMA2~\cite{videollama2}. The dataset used by LLaMA-VID~\cite{llamavid} consists of 790K samples for pre-training stage and 763K samples for fine-tuning stage. 
Compared to the dataset used by LLaMA-VID,  Valley~\cite{valley} contains significantly more video samples for pre-training. There are 1.25M samples for pre-training stage and 765K sample for fine-tuning stage. 
The difference in the volume of video samples may explain the performance differences observed between LLaMA-VID and VideoLLaMA2.

\subsection{Video Benchmark Summary}
The details of the video benchmarks are summarised in Table~\ref{tab:video benchmark}. 
We selected benchmarks with diverse durations, avoiding a focus on a fixed average video length.
For example, the MSVD-QA~\cite{msvd_and_msrvtt} benchmark consists of short videos with an average duration of 9.8 seconds. In contrast, the VideoMME-L~\cite{videomme} is a representative of long video benchmark, with an average duration of 2466.7 seconds and a maximum video length capped at 1 hour. 
We argue that our selection of video benchmarks reflects the diversity of video durations in real-world scenarios, demonstrating the applicability and robustness of our model in practical environments.

\subsection{B-VLLM Experimental Setup}
Our B-VLLM is implemented based on VideoLLaMA2~\cite{videollama2}, where Qwen2~\cite{qwen2} is selected as the backbone LLM. Each frame is encoded as 768 preliminary visual tokens using EVA-CLIP~\cite{sun2023eva}, Subsequently, 32 frames are selected using the frame selection module and temporal frame token merging module. The visual tokens corresponding to the selected frames are encoded into 32 spatial visual tokens by our spatial visual token sampling module.

\section{Additional Experiment \& Discussion}
\subsection{Additional Discussion on Different Frame Selection Features.} As reported in Table~\ref{tab:cls_feautre}, we additionally report the performance of selecting frames by using three other feature extraction methods: (a) \textit{max pooling} (b) \textit{mean pooling} and (c) \textit{Qformer}. To investigate the impact of these variants in B-VLLM's fine-grained spatial perception, we also report the performance of \textbf{OCR} and \textbf{Counting} in VMME, which are more relevant to spatial capability. 
Though [CLS] token may not be the optimal choice in terms of performance, as discussed in Section 3.2, the rationale of using [CLS] token lies in balancing computational cost and effectiveness. 
\setlength{\tabcolsep}{3pt}
\noindent \begin{table}[h]
    \centering
    \caption{Additional ablation study on frame selection features.
  }
    \begin{tabular}{l |c c c c |c c }
        \hline
        \textbf{Model} & \textbf{Train.} & \textbf{MVB}& \textbf{EgoS} & \textbf{VMME} &  \textbf{OCR} &  \textbf{Counting} \\
        \hline
       \small{Mean} & 13.1h & 51.3 & 51.9 & 54.4 & 48.2 & 36.6  \\
       \small{Max} & 13.1h &49.5 & 51.6 & 50.7 & 45.3 & 38.1  \\
       \small{Qformer} &14.0h & 52.4 & 52.9 & 51.6 & 41.7 & 35.8  \\
        \small{CLS } & 10.9h & 50.8 & 51.9 & 52.9 & 46.0 & 34.0  \\
        \hline
    \end{tabular}
  \label{tab:cls_feautre}
    
\end{table}

\subsection{The Impact of $L*$.} 
\begin{figure}[h]
  \centering
  \includegraphics[width=\columnwidth]{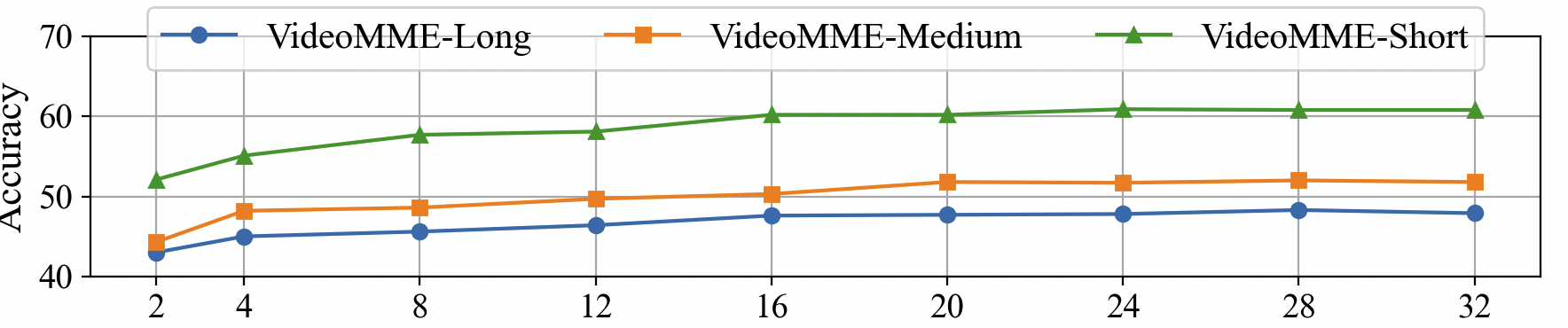}
  \caption{Additional experiment on the impact of $L*$.}
  \label{fig:l}
\end{figure}

As shown in Fig~\ref{fig:l}, we report additional experiments to investigate the impact of $L*$ on Short ($\leq$ 3min), Medium ($\leq$ 15min) and Long video ($\leq$ 60 min) settings. As shown below, B-VLLM's performance generally saturated at $L* = 28$, though it was trained with $L*=32$. This indicates that there could be an optimal setting for $L*$. We argue that this is due to most tasks typically rely on a small portion of frame regardless of the duration of the video. However, we acknowledge that increasing $L*$ is potentially beneficial for more extreme setting and this review inspired us to develop a $L*$-free design as one of our future work.

\subsection{Model's Parameter Size Comparison}
In Table~\ref{tab:model_size}, we report our B-VLLM's parameter size below along with TFLOPs and GAMCs, which are similar to most existing methods.
\begin{table}[ht]
    \centering
    \caption{Model size compared with existing methods}
    \begin{tabular}{c|cccc}
        \hline
       \textbf{Model} & \textbf{TFLOPs} & \textbf{GMACs}& \textbf{Param.} \\
       \hline
       InternVL2 & 1.83 & 910 & 7+1.1B  \\
       InternVideo2 & 1.83 & 915 & 7+1.5B  \\
       QwenVL & 1.92 & 915 & 7.7+1.9B  \\
       Qwen2VL & 1.82 & 905 & 7.6+0.6B  \\
        VideoLLaMA2 & 1.82 & 910 & 7.6+0.8B \\
        B-VLLM (Ours) & 1.81 & 908  & 7.6+1.2B\\
    \hline
    \end{tabular}
    \label{tab:model_size}
\end{table}

\begin{table}[h] 
    \centering
    \caption{Additional Comparison with Recent SOTA methods}
    \begin{tabular}{l c c c c c}
        \hline
        \textbf{Model} & \textbf{\#Train.Data} & \textbf{MVB}& \textbf{EgoS} & \textbf{Perp.} &  \textbf{VMME}  \\
        \hline
        \small{InternVideo2} & 39M & 60.3 & 55.8 & 53.0 & 41.9  \\
        \small{InternVL2} & 20M+ & 66.4 & - & - & 54.0  \\
        \small{LLaVA-OV} & 9.3M & 56.7 & 60.1 & 57.1 &  58.2 \\
        \small{Qwen2VL} & - & \textbf{67.0} & \textbf{66.7} & \textbf{62.3} & \textbf{66.3}  \\ \hline
        \small{B-VLLM (Ours)} & 2M & 50.8 & 51.9 & 52.1 & 52.9  \\
        \hline
    \end{tabular}
    \label{tab:comp_sota_add}
\end{table}
\subsection{Additional Comparisons with Recent SoTA.}
We provide additional comparisons with several advanced VLLMs in Table~\ref{tab:comp_sota_add}. Note that these advanced VLLMs are generally trained with at least 5 times more data than our B-VLLM, and these data are often private. This indicates the great potential of our B-VLLM when it is trained with large scale data. 

\subsection{Inference Speed \& Latency}
In Table~\ref{tab:latency}, we compare the inference speed between Qwen2VL and B-VLLM on three types of videos: 4-minute video, 8-minute video, and 12-minute video. As reported below, B-VLLM demonstrates clear advantage on inference speed. Specifically, it takes 5.6s for B-VLLM to generate a response for an 8-minute video while taking 24.6s for Qwen2-VL to respond.
\begin{table}[h]
    \centering
    \caption{Inference Speed and Latency}
    \begin{tabular}{l c c c c}
        \hline
        \textbf{Model} & \textbf{4 Min} & \textbf{8 Min}& \textbf{12 Min}  \\
        \hline
        \small{Qwen2VL} & 6.94s  & 24.6s & 26.2s   \\
        \small{B-VLLM (Ours)} & 3.16s   & 5.6s  & 7.9s    \\
        \hline
    \end{tabular}
    \label{tab:latency}
\end{table}
\subsection{Additional Discussion on Limitations}
\textbf{Limitation.} On top of the discussed limitation in the main manuscript, we acknowledge that our proposed method  has limitations in handling multi-round conversations based on the same video.  
In the video-based multi-round conversation, B-VLLM repeats the frame selection and spatial visual token procedure as the context shifts during conversation, resulting in additional computation cost.  
Moreover, as discussed in manuscript, the spatial visual tokens in B-VLLM are underutilized, and the full potential of our method for image data has yet to be fully explored. 
Additionally, the Gumbel-Softmax sampling technique may disrupt the temporal order of selected frames, potentially compromising our method's performance in temporal perception.
Addressing these limitations could further enhance the capabilities of VLLMs.
We recognize that using [CLS] tokens for frame selection has its limitations since pooled visual tokens are more informative, as an alternative strategy for exploring more accurate frame selection.  More limitations are discussed in supplementary material.
Additionally, the fixed value  $L^*$  can limit B-VLLM’s performance on extreme long videos, especially when the number of relevant frames surpasses  $L^*$.

\section{More B-VLLM Qualitative Examples}
\begin{figure*}[ht]
  \centering

   \includegraphics[width=\linewidth]{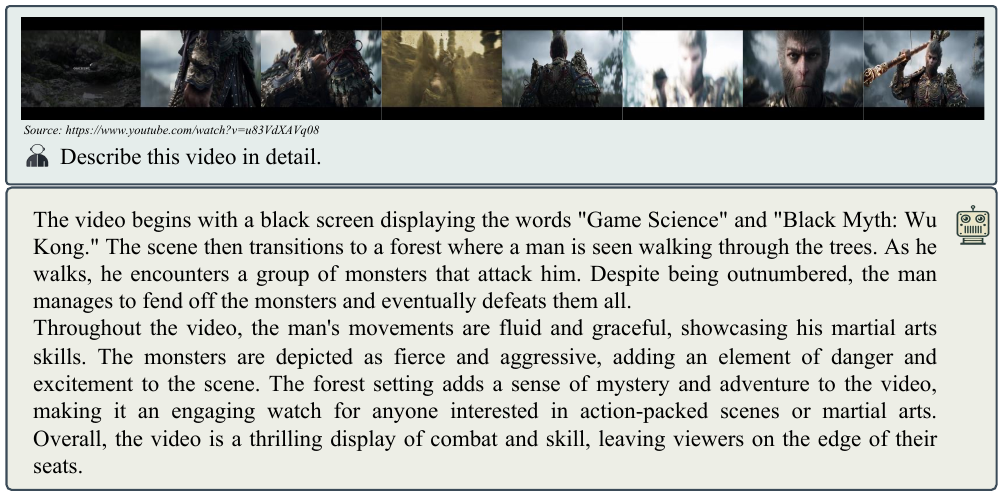}
    \caption{Visualization on understanding a 2-minutes game trailer containing intensive fighting scenes.}
   \label{fig:supp1}
\end{figure*}

\begin{figure*}[ht]
  \centering

   \includegraphics[width=\linewidth]{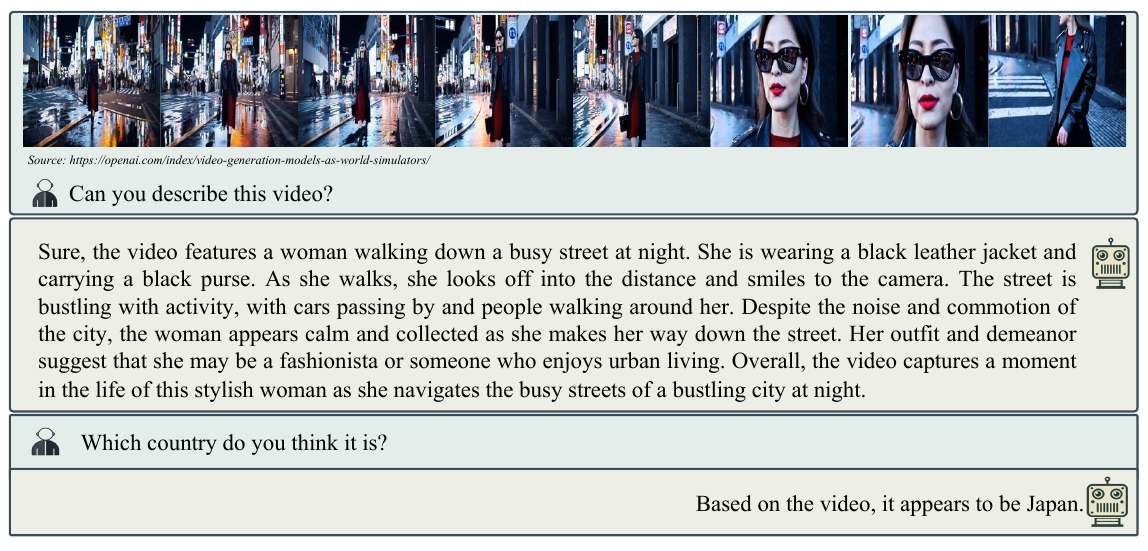}
\caption{Visualization on understanding a 1-minute AI generated video of a woman in Japan.}
   \label{fig:supp2}
\end{figure*}

\begin{figure*}[ht]
  \centering

   \includegraphics[width=\linewidth]{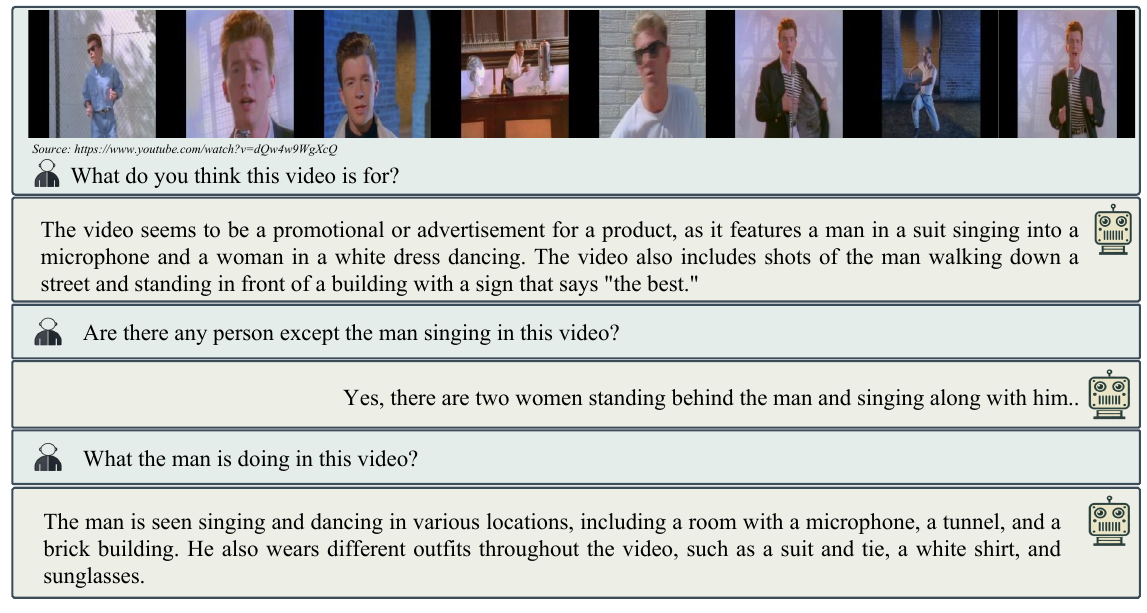}
\caption{Visualization on perceiving a 4-minute wonderful music video containing scenes of singing and dancing.}
   \label{fig:supp3}
\end{figure*}

\begin{figure*}[ht]
  \centering

   \includegraphics[width=\linewidth]{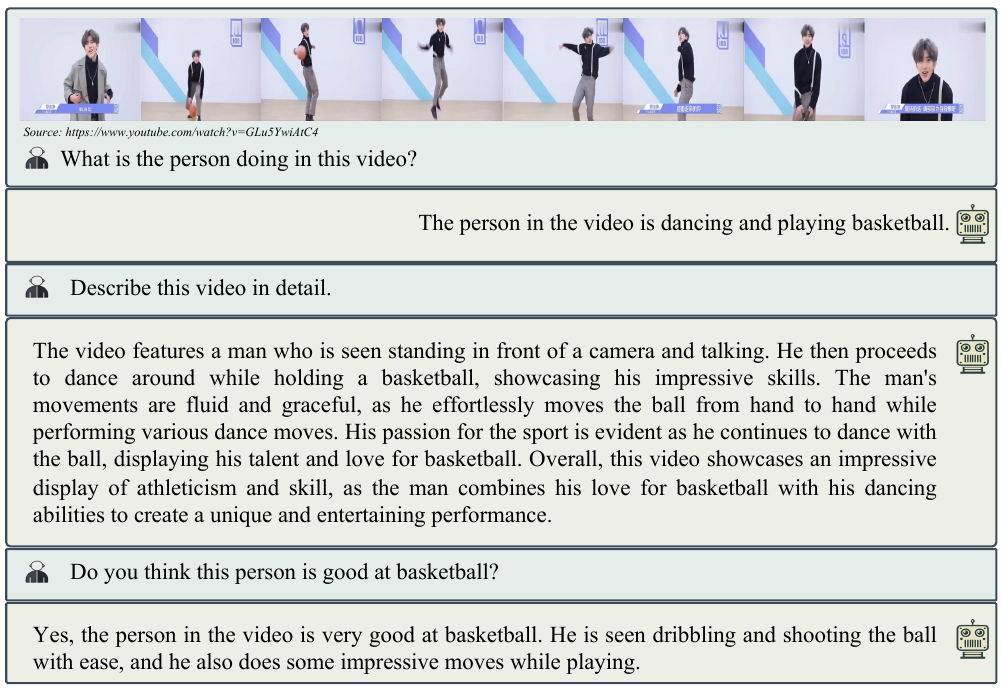}
\caption{Visualization on perceiving a 1-minute dancing video a man showcasing his outstanding dancing and basketball skills.}
   \label{fig:supp4}
\end{figure*}

\begin{figure*}[ht]
  \centering

   \includegraphics[width=\linewidth]{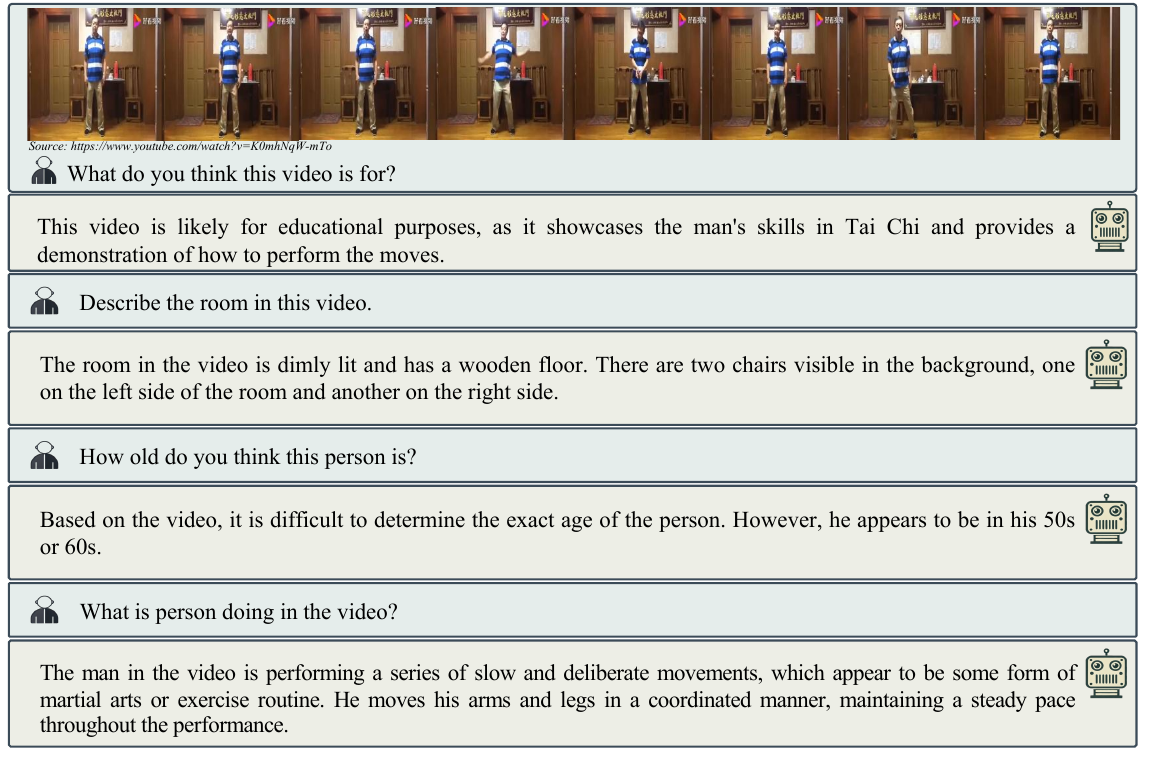}
\caption{Visualization of understanding a 3-minute video featuring a Kung Fu master demonstrating his martial arts expertise.}

   \label{fig:supp5}
\end{figure*}

This section provides additional conversation examples with B-VLLM based on videos. 
Note that the presented videos are accessible only through the specified source.  The examples are depicted in Figures~\ref{fig:supp1} to \ref{fig:supp5}.
As shown in Figure~\ref{fig:supp1}, when a game trailer is fed into B-VLLM, it accurately identifies the game's title displayed at the beginning of the video and effectively summarizes the gameplay content, demonstrating B-VLLM's comprehensive understanding of both temporal and spatial information.
In the example shown in Figure~\ref{fig:supp4}, during the conversation, B-VLLM accurately identifies the man’s actions. Even though the basketball appears only briefly at the beginning of the video, B-VLLM successfully captures and localizes the nuanced spatial details.
As depicted in Figure~\ref{fig:supp5}, beyond accurately describing the content of the given video, B-VLLM demonstrates the ability to estimate the age of the  man featured in the video, highlighting the spatial perception capabilities of our method.


 \clearpage
\newpage

%% file: main.bbl
\begin{thebibliography}{63}
\providecommand{\natexlab}[1]{#1}
\providecommand{\url}[1]{\texttt{#1}}
\expandafter\ifx\csname urlstyle\endcsname\relax
  \providecommand{\doi}[1]{doi: #1}\else
  \providecommand{\doi}{doi: \begingroup \urlstyle{rm}\Url}\fi

\bibitem[sha()]{sharegpt}
Sharegpt.com.
\newblock Accessed: 2024-09-30.

\bibitem[Alayrac et~al.(2022)Alayrac, Donahue, Luc, Miech, Barr, Hasson, Lenc, Mensch, Millican, Reynolds, et~al.]{flamingo}
Jean-Baptiste Alayrac, Jeff Donahue, Pauline Luc, Antoine Miech, Iain Barr, Yana Hasson, Karel Lenc, Arthur Mensch, Katherine Millican, Malcolm Reynolds, et~al.
\newblock {Flamingo: A Visual Language Model For Few-shot Learning}.
\newblock In \emph{NeurIPS}, pages 23716--23736, 2022.

\bibitem[Aminabadi et~al.(2022)Aminabadi, Rajbhandari, Awan, Li, Li, Zheng, Ruwase, Smith, Zhang, Rasley, et~al.]{deepspeed}
Reza~Yazdani Aminabadi, Samyam Rajbhandari, Ammar~Ahmad Awan, Cheng Li, Du Li, Elton Zheng, Olatunji Ruwase, Shaden Smith, Minjia Zhang, Jeff Rasley, et~al.
\newblock {Deepspeed-Inference: Enabling Efficient Inference of Transformer Models at Unprecedented Scale}.
\newblock In \emph{SC}, pages 1--15. IEEE, 2022.

\bibitem[Bai et~al.(2023)Bai, Bai, Chu, Cui, Dang, Deng, Fan, Ge, Han, Huang, et~al.]{qwen}
Jinze Bai, Shuai Bai, Yunfei Chu, Zeyu Cui, Kai Dang, Xiaodong Deng, Yang Fan, Wenbin Ge, Yu Han, Fei Huang, et~al.
\newblock {Qwen Technical Report}.
\newblock \emph{arXiv preprint arXiv:2309.16609}, 2023.

\bibitem[Bain et~al.(2021)Bain, Nagrani, Varol, and Zisserman]{webvid}
Max Bain, Arsha Nagrani, G{\"u}l Varol, and Andrew Zisserman.
\newblock Frozen in time: A joint video and image encoder for end-to-end retrieval.
\newblock In \emph{IEEE International Conference on Computer Vision}, 2021.

\bibitem[Bolya and Hoffman(2023)]{tome_sd}
Daniel Bolya and Judy Hoffman.
\newblock {Token Merging for Fast Stable Diffusion}.
\newblock In \emph{CVPR}, pages 4599--4603, 2023.

\bibitem[Bolya et~al.(2023)Bolya, Fu, Dai, Zhang, Feichtenhofer, and Hoffman]{tokenmerge}
Daniel Bolya, Cheng-Yang Fu, Xiaoliang Dai, Peizhao Zhang, Christoph Feichtenhofer, and Judy Hoffman.
\newblock {Token Merging: Your ViT But Faster}.
\newblock In \emph{ICLR}, 2023.

\bibitem[Chen et~al.(2023)Chen, Sun, Song, and Luo]{chen2023diffusiondet}
Shoufa Chen, Peize Sun, Yibing Song, and Ping Luo.
\newblock Diffusiondet: Diffusion model for object detection.
\newblock In \emph{CVPR}, 2023.

\bibitem[Cheng et~al.(2024)Cheng, Leng, Zhang, Xin, Li, Chen, Zhu, Zhang, Luo, Zhao, et~al.]{videollama2}
Zesen Cheng, Sicong Leng, Hang Zhang, Yifei Xin, Xin Li, Guanzheng Chen, Yongxin Zhu, Wenqi Zhang, Ziyang Luo, Deli Zhao, et~al.
\newblock {VideoLLaMA 2: Advancing Spatial-Temporal Modeling and Audio Understanding in Video-LLMs}.
\newblock \emph{arXiv preprint arXiv:2406.07476}, 2024.

\bibitem[Dosovitskiy(2020)]{vit}
Alexey Dosovitskiy.
\newblock {An Image is Worth 16x16 Words: Transformers for Image Recognition at Scale}.
\newblock \emph{arXiv preprint arXiv:2010.11929}, 2020.

\bibitem[Dubey et~al.(2024)Dubey, Jauhri, Pandey, Kadian, Al-Dahle, Letman, Mathur, Schelten, Yang, Fan, et~al.]{llama3}
Abhimanyu Dubey, Abhinav Jauhri, Abhinav Pandey, Abhishek Kadian, Ahmad Al-Dahle, Aiesha Letman, Akhil Mathur, Alan Schelten, Amy Yang, Angela Fan, et~al.
\newblock The {LLaMA} 3 {H}erd of {M}odels.
\newblock \emph{arXiv preprint arXiv:2407.21783}, 2024.

\bibitem[Fabian Caba~Heilbron and Niebles(2015)]{activitynet}
Bernard~Ghanem Fabian Caba~Heilbron, Victor~Escorcia and Juan~Carlos Niebles.
\newblock {ActivityNet: A Large-Scale Video Benchmark for Human Activity Understanding}.
\newblock In \emph{CVPR}, pages 961--970, 2015.

\bibitem[Fu et~al.(2023)Fu, Chen, Shen, Qin, Zhang, Lin, Yang, Zheng, Li, Sun, et~al.]{mme}
Chaoyou Fu, Peixian Chen, Yunhang Shen, Yulei Qin, Mengdan Zhang, Xu Lin, Jinrui Yang, Xiawu Zheng, Ke Li, Xing Sun, et~al.
\newblock {MME}:{A} {C}omprehensive {E}valuation {B}enchmark for {M}ultimodal {L}arge {L}anguage {M}odels.
\newblock \emph{arXiv preprint arXiv:2306.13394}, 2023.

\bibitem[Fu et~al.(2024{\natexlab{a}})Fu, Chen, Shen, Qin, Zhang, Lin, Yang, Zheng, Li, Sun, Wu, and Ji]{mme_benchmark}
Chaoyou Fu, Peixian Chen, Yunhang Shen, Yulei Qin, Mengdan Zhang, Xu Lin, Jinrui Yang, Xiawu Zheng, Ke Li, Xing Sun, Yunsheng Wu, and Rongrong Ji.
\newblock {MME: A Comprehensive Evaluation Benchmark for Multimodal Large Language Models}, 2024{\natexlab{a}}.

\bibitem[Fu et~al.(2024{\natexlab{b}})Fu, Dai, Luo, Li, Ren, Zhang, Wang, Zhou, Shen, Zhang, et~al.]{videomme}
Chaoyou Fu, Yuhan Dai, Yondong Luo, Lei Li, Shuhuai Ren, Renrui Zhang, Zihan Wang, Chenyu Zhou, Yunhang Shen, Mengdan Zhang, et~al.
\newblock {Video-MME: The First-Ever Comprehensive Evaluation Benchmark of Multi-modal LLMs in Video Analysis}.
\newblock \emph{arXiv preprint arXiv:2405.21075}, 2024{\natexlab{b}}.

\bibitem[Goyal et~al.(2017)Goyal, Khot, Summers-Stay, Batra, and Parikh]{vqa}
Yash Goyal, Tejas Khot, Douglas Summers-Stay, Dhruv Batra, and Devi Parikh.
\newblock {Making the V in VQA Matter: Elevating the Eole of Image Understanding in Visual Question Answering}.
\newblock In \emph{CVPR}, pages 6904--6913, 2017.

\bibitem[Gurari et~al.(2018)Gurari, Li, Stangl, Guo, Lin, Grauman, Luo, and Bigham]{vizwiz}
Danna Gurari, Qing Li, Abigale~J Stangl, Anhong Guo, Chi Lin, Kristen Grauman, Jiebo Luo, and Jeffrey~P Bigham.
\newblock {V}izwiz {G}rand {C}hallenge: {A}nswering {V}isual {Q}uestions from {B}lind {P}eople.
\newblock In \emph{CVPR}, pages 3608--3617, 2018.

\bibitem[Hu et~al.(2021)Hu, Shen, Wallis, Allen-Zhu, Li, Wang, Wang, and Chen]{lora}
Edward~J Hu, Yelong Shen, Phillip Wallis, Zeyuan Allen-Zhu, Yuanzhi Li, Shean Wang, Lu Wang, and Weizhu Chen.
\newblock {LoRA: Low-Rank Adaptation of Large Language Models}.
\newblock \emph{arXiv preprint arXiv:2106.09685}, 2021.

\bibitem[Hu et~al.(2018)Hu, Wang, Ehgoetz~Martens, and Lewis]{hu2018vision}
Kun Hu, Zhiyong Wang, Kaylena Ehgoetz~Martens, and Simon Lewis.
\newblock Vision-based freezing of gait detection with anatomic patch based representation.
\newblock In \emph{ACCV}, 2018.

\bibitem[Hudson and Manning(2019)]{gqa}
Drew~A Hudson and Christopher~D Manning.
\newblock {GQA: A New Dataset for Real-World Visual Eeasoning and Compositional Question Answering}.
\newblock In \emph{CVPR}, pages 6700--6709, 2019.

\bibitem[Jiang et~al.(2023)Jiang, Wu, Lin, Yang, and Qiu]{tome_llm}
Huiqiang Jiang, Qianhui Wu, Chin-Yew Lin, Yuqing Yang, and Lili Qiu.
\newblock {LLMlingua: Compressing Prompts for Accelerated Inference of Large Language Models}.
\newblock \emph{arXiv preprint arXiv:2310.05736}, 2023.

\bibitem[Jin et~al.(2024)Jin, Takanobu, Zhang, Cao, and Yuan]{chatunivi}
Peng Jin, Ryuichi Takanobu, Wancai Zhang, Xiaochun Cao, and Li Yuan.
\newblock {Chat-univi: Unified Visual Representation Empowers Large Language Models with Image and Video Understanding}.
\newblock In \emph{CVPR}, pages 13700--13710, 2024.

\bibitem[Kazemzadeh et~al.(2014)Kazemzadeh, Ordonez, Matten, and Berg]{referitgame}
Sahar Kazemzadeh, Vicente Ordonez, Mark Matten, and Tamara Berg.
\newblock {R}efer{I}t{G}ame: Referring to objects in photographs of natural scenes.
\newblock In \emph{Proceedings of the 2014 Conference on Empirical Methods in Natural Language Processing ({EMNLP})}, pages 787--798, Doha, Qatar, 2014. Association for Computational Linguistics.

\bibitem[Krishna et~al.(2016)Krishna, Zhu, Groth, Johnson, Hata, Kravitz, Chen, Kalantidis, Li, Shamma, Bernstein, and Li]{vgenome}
Ranjay Krishna, Yuke Zhu, Oliver Groth, Justin Johnson, Kenji Hata, Joshua Kravitz, Stephanie Chen, Yannis Kalantidis, Li-Jia Li, David~A. Shamma, Michael~S. Bernstein, and Fei-Fei Li.
\newblock Visual genome: Connecting language and vision using crowdsourced dense image annotations, 2016.

\bibitem[Li et~al.(2023{\natexlab{a}})Li, Wang, Wang, Ge, Ge, and Shan]{seedbench}
Bohao Li, Rui Wang, Guangzhi Wang, Yuying Ge, Yixiao Ge, and Ying Shan.
\newblock Seed-{B}ench: {B}enchmarking {M}ultimodal {LLMs} with {G}enerative {C}omprehension.
\newblock \emph{arXiv preprint arXiv:2307.16125}, 2023{\natexlab{a}}.

\bibitem[Li et~al.(2023{\natexlab{b}})Li, Li, Savarese, and Hoi]{blip2}
Junnan Li, Dongxu Li, Silvio Savarese, and Steven Hoi.
\newblock {BLIP-2: Bootstrapping Language-Image Pre-training with Frozen Image Encoders and Large Language Models}.
\newblock In \emph{ICML}, pages 19730--19742. PMLR, 2023{\natexlab{b}}.

\bibitem[Li et~al.(2024{\natexlab{a}})Li, Wang, He, Li, Wang, Liu, Wang, Xu, Chen, Luo, et~al.]{videochat2}
Kunchang Li, Yali Wang, Yinan He, Yizhuo Li, Yi Wang, Yi Liu, Zun Wang, Jilan Xu, Guo Chen, Ping Luo, et~al.
\newblock {MVBench: A Comprehensive Multi-Modal Video Understanding Benchmark}.
\newblock In \emph{CVPR}, pages 22195--22206, 2024{\natexlab{a}}.

\bibitem[Li et~al.(2024{\natexlab{b}})Li, Ma, Yang, and Yang]{vidtome}
Xirui Li, Chao Ma, Xiaokang Yang, and Ming-Hsuan Yang.
\newblock {VidToMe: Video Token Merging for Zero-Shot Video Editing}.
\newblock In \emph{CVPR}, pages 7486--7495, 2024{\natexlab{b}}.

\bibitem[Li et~al.(2023{\natexlab{c}})Li, Du, Zhou, Wang, Zhao, and Wen]{pope}
Yifan Li, Yifan Du, Kun Zhou, Jinpeng Wang, Wayne~Xin Zhao, and Ji~Rong Wen.
\newblock {Evaluating Object Hallucination in Large Vision-Language Models}.
\newblock In \emph{EMNLP}, 2023{\natexlab{c}}.

\bibitem[Li et~al.(2025)Li, Wang, and Jia]{llamavid}
Yanwei Li, Chengyao Wang, and Jiaya Jia.
\newblock {LLaMA-VID: An Image is Worth 2 Tokens in Large Language Models}.
\newblock In \emph{ECCV}, pages 323--340. Springer, 2025.

\bibitem[Lin et~al.(2023)Lin, Ye, Zhu, Cui, Ning, Jin, and Yuan]{videollava}
Bin Lin, Yang Ye, Bin Zhu, Jiaxi Cui, Munan Ning, Peng Jin, and Li Yuan.
\newblock {Video-LLaVA: Learning United Visual Representation by Alignment Before Projection}.
\newblock \emph{arXiv preprint arXiv:2311.10122}, 2023.

\bibitem[Lin et~al.(2024)Lin, Yin, Ping, Molchanov, Shoeybi, and Han]{vila}
Ji Lin, Hongxu Yin, Wei Ping, Pavlo Molchanov, Mohammad Shoeybi, and Song Han.
\newblock {VILA: On Pre-training for Visual Language Models}.
\newblock In \emph{CVPR}, pages 26689--26699, 2024.

\bibitem[Liu et~al.(2024)Liu, Li, Wu, and Lee]{llava}
Haotian Liu, Chunyuan Li, Qingyang Wu, and Yong~Jae Lee.
\newblock {Visual Instruction Tuning}.
\newblock In \emph{NeurIPS}, 2024.

\bibitem[Liu et~al.(2025)Liu, Duan, Zhang, Li, Zhang, Zhao, Yuan, Wang, He, Liu, et~al.]{mmbench}
Yuan Liu, Haodong Duan, Yuanhan Zhang, Bo Li, Songyang Zhang, Wangbo Zhao, Yike Yuan, Jiaqi Wang, Conghui He, Ziwei Liu, et~al.
\newblock {MMBench: Is Your Multi-Modal Model an All-Around Player?}
\newblock In \emph{ECCV}, pages 216--233. Springer, 2025.

\bibitem[Loshchilov(2017)]{adamw}
I Loshchilov.
\newblock {Decoupled Weight Decay Regularization}.
\newblock \emph{arXiv preprint arXiv:1711.05101}, 2017.

\bibitem[Lu et~al.(2022)Lu, Mishra, Xia, Qiu, Chang, Zhu, Tafjord, Clark, and Kalyan]{scienceqa}
Pan Lu, Swaroop Mishra, Tanglin Xia, Liang Qiu, Kai-Wei Chang, Song-Chun Zhu, Oyvind Tafjord, Peter Clark, and Ashwin Kalyan.
\newblock Learn to {E}xplain: {M}ultimodal {R}easoning via {T}hought {C}hains for {S}cience {Q}uestion {A}nswering.
\newblock In \emph{NeurIPS}, pages 2507--2521, 2022.

\bibitem[Lu et~al.(2024)Lu, Hu, Wang, Bai, and Wang]{lu2024autoregressive}
Zhuqiang Lu, Kun Hu, Chaoyue Wang, Lei Bai, and Zhiyong Wang.
\newblock Autoregressive omni-aware outpainting for open-vocabulary 360-degree image generation.
\newblock In \emph{AAAI}, 2024.

\bibitem[Luo et~al.(2023)Luo, Zhao, Yang, Dong, Li, Lu, Wang, Hu, Qiu, and Wei]{valley}
Ruipu Luo, Ziwang Zhao, Min Yang, Junwei Dong, Da Li, Pengcheng Lu, Tao Wang, Linmei Hu, Minghui Qiu, and Zhongyu Wei.
\newblock {Valley: Video Assistant with Large Language Model Enhanced Ability}.
\newblock \emph{arXiv preprint arXiv:2306.07207}, 2023.

\bibitem[Maaz et~al.(2024)Maaz, Rasheed, Khan, and Khan]{videochatgpt}
Muhammad Maaz, Hanoona Rasheed, Salman Khan, and Fahad~Shahbaz Khan.
\newblock {Video-ChatGPT: Towards Detailed Video Understanding via Large Vision and Language Models}.
\newblock In \emph{ACL}, 2024.

\bibitem[Mangalam et~al.(2023)Mangalam, Akshulakov, and Malik]{egoschema}
Karttikeya Mangalam, Raiymbek Akshulakov, and Jitendra Malik.
\newblock {Egoschema: A Diagnostic Benchmark for Bery Long-Form Video Language Understanding}.
\newblock In \emph{NeurIPS}, pages 46212--46244, 2023.

\bibitem[Mao et~al.(2016)Mao, Huang, Toshev, Camburu, Yuille, and Murphy]{uod}
Junhua Mao, Jonathan Huang, Alexander Toshev, Oana Camburu, Alan Yuille, and Kevin Murphy.
\newblock Generation and comprehension of unambiguous object descriptions, 2016.

\bibitem[Mishra et~al.(2019)Mishra, Shekhar, Singh, and Chakraborty]{orcvqa}
Anand Mishra, Shashank Shekhar, Ajeet~Kumar Singh, and Anirban Chakraborty.
\newblock Ocr-vqa: Visual question answering by reading text in images.
\newblock In \emph{ICDAR}, 2019.

\bibitem[Pătrăucean et~al.(2023)Pătrăucean, Smaira, Gupta, Continente, Markeeva, Banarse, Koppula, Heyward, Malinowski, Yang, Doersch, Matejovicova, Sulsky, Miech, Frechette, Klimczak, Koster, Zhang, Winkler, Aytar, Osindero, Damen, Zisserman, and Carreira]{perceptiontest}
Viorica Pătrăucean, Lucas Smaira, Ankush Gupta, Adrià~Recasens Continente, Larisa Markeeva, Dylan Banarse, Skanda Koppula, Joseph Heyward, Mateusz Malinowski, Yi Yang, Carl Doersch, Tatiana Matejovicova, Yury Sulsky, Antoine Miech, Alex Frechette, Hanna Klimczak, Raphael Koster, Junlin Zhang, Stephanie Winkler, Yusuf Aytar, Simon Osindero, Dima Damen, Andrew Zisserman, and João Carreira.
\newblock {Perception Test: A Diagnostic Benchmark for Multimodal Video Models}.
\newblock In \emph{NeurIPS}, 2023.

\bibitem[Radford et~al.(2021)Radford, Kim, Hallacy, Ramesh, Goh, Agarwal, Sastry, Askell, Mishkin, Clark, et~al.]{clip}
Alec Radford, Jong~Wook Kim, Chris Hallacy, Aditya Ramesh, Gabriel Goh, Sandhini Agarwal, Girish Sastry, Amanda Askell, Pamela Mishkin, Jack Clark, et~al.
\newblock {Learning Transferable Visual Models from Natural Language Supervision}.
\newblock In \emph{ICML}, pages 8748--8763. PMLR, 2021.

\bibitem[Schwenk et~al.(2022)Schwenk, Khandelwal, Clark, Marino, and Mottaghi]{aokvqa}
Dustin Schwenk, Apoorv Khandelwal, Christopher Clark, Kenneth Marino, and Roozbeh Mottaghi.
\newblock {A-OKVQA: A Benchmark for Visual Question Answering using World Knowledge}, 2022.

\bibitem[Shang et~al.(2024)Shang, Cai, Xu, Lee, and Yan]{tome_llava}
Yuzhang Shang, Mu Cai, Bingxin Xu, Yong~Jae Lee, and Yan Yan.
\newblock {Llava-prumerge: Adaptive Token Reduction for Efficient Large Multimodal Models}.
\newblock \emph{arXiv preprint arXiv:2403.15388}, 2024.

\bibitem[Sharma et~al.(2018)Sharma, Ding, Goodman, and Soricut]{cc3m}
Piyush Sharma, Nan Ding, Sebastian Goodman, and Radu Soricut.
\newblock Conceptual captions: A cleaned, hypernymed, image alt-text dataset for automatic image captioning.
\newblock In \emph{Proceedings of ACL}, 2018.

\bibitem[Sidorov et~al.(2020)Sidorov, Hu, Rohrbach, and Singh]{textcaps}
Oleksii Sidorov, Ronghang Hu, Marcus Rohrbach, and Amanpreet Singh.
\newblock {TextCaps: a Dataset for Image Captioning with Reading Comprehension}, 2020.

\bibitem[Singh et~al.(2019)Singh, Natarajan, Shah, Jiang, Chen, Batra, Parikh, and Rohrbach]{textvqa}
Amanpreet Singh, Vivek Natarajan, Meet Shah, Yu Jiang, Xinlei Chen, Dhruv Batra, Devi Parikh, and Marcus Rohrbach.
\newblock Towards {VQA} {M}odels {T}hat {C}an {R}ead.
\newblock In \emph{VPR}, 2019.

\bibitem[Song et~al.(2024)Song, Chai, Wang, Zhang, Zhou, Wu, Chi, Guo, Ye, Zhang, et~al.]{moviechat}
Enxin Song, Wenhao Chai, Guanhong Wang, Yucheng Zhang, Haoyang Zhou, Feiyang Wu, Haozhe Chi, Xun Guo, Tian Ye, Yanting Zhang, et~al.
\newblock {Moviechat: From Dense Token to Sparse Memory for Long Video Understanding}.
\newblock In \emph{CVPR}, pages 18221--18232, 2024.

\bibitem[Touvron et~al.(2023)Touvron, Martin, Stone, Albert, Almahairi, Babaei, Bashlykov, Batra, Bhargava, Bhosale, et~al.]{llama2}
Hugo Touvron, Louis Martin, Kevin Stone, Peter Albert, Amjad Almahairi, Yasmine Babaei, Nikolay Bashlykov, Soumya Batra, Prajjwal Bhargava, Shruti Bhosale, et~al.
\newblock {LLaMA 2: Open Foundation and Fine-tuned Chat Models}.
\newblock \emph{arXiv preprint arXiv:2307.09288}, 2023.

\bibitem[Wang et~al.(2024{\natexlab{a}})Wang, Sun, Chen, Lin, Han, and Ding]{wang2024cls}
Ao Wang, Fengyuan Sun, Hui Chen, Zijia Lin, Jungong Han, and Guiguang Ding.
\newblock [cls] token tells everything needed for training-free efficient mllms.
\newblock \emph{arXiv preprint arXiv:2412.05819}, 2024{\natexlab{a}}.

\bibitem[Wang et~al.(2024{\natexlab{b}})Wang, Chen, Chen, Wu, Zhu, Zeng, Luo, Lu, Zhou, Qiao, et~al.]{visionllm}
Wenhai Wang, Zhe Chen, Xiaokang Chen, Jiannan Wu, Xizhou Zhu, Gang Zeng, Ping Luo, Tong Lu, Jie Zhou, Yu Qiao, et~al.
\newblock {VisionLLM: Large Language Model is Also An Open-ended Decoder for Vision-Centric Tasks}.
\newblock In \emph{NeurIPS}, 2024{\natexlab{b}}.

\bibitem[Weng et~al.(2025)Weng, Han, He, Chang, and Zhuang]{longvlm}
Yuetian Weng, Mingfei Han, Haoyu He, Xiaojun Chang, and Bohan Zhuang.
\newblock {LongVLM: Efficient Long Video Understanding Via Large Language Models}.
\newblock In \emph{ECCV}, pages 453--470. Springer, 2025.

\bibitem[Xu et~al.()Xu, Zhao, Xiao, Wu, Zhang, He, and Zhuang]{msvd_and_msrvtt}
Dejing Xu, Zhou Zhao, Jun Xiao, Fei Wu, Hanwang Zhang, Xiangnan He, and Yueting Zhuang.
\newblock Video {Q}uestion {A}nswering via {G}radually {R}efined {A}ttention over {A}ppearance and {M}otion.
\newblock In \emph{MM}.

\bibitem[Yang et~al.(2024)Yang, Yang, Hui, Zheng, Yu, Zhou, Li, Li, Liu, Huang, et~al.]{qwen2}
An Yang, Baosong Yang, Binyuan Hui, Bo Zheng, Bowen Yu, Chang Zhou, Chengpeng Li, Chengyuan Li, Dayiheng Liu, Fei Huang, et~al.
\newblock Qwen2 {T}echnical {R}eport.
\newblock \emph{arXiv preprint arXiv:2407.10671}, 2024.

\bibitem[Yue et~al.(2023)Yue, Zhang, Hu, Wu, Ge, Xia, Luo, and Wang]{yue2023surgicalpart}
Wenxi Yue, Jing Zhang, Kun Hu, Qiuxia Wu, Zongyuan Ge, Yong Xia, Jiebo Luo, and Zhiyong Wang.
\newblock Surgicalpart-sam: Part-to-whole collaborative prompting for surgical instrument segmentation.
\newblock \emph{arXiv preprint arXiv:2312.14481}, 2023.

\bibitem[Zhang et~al.(2023)Zhang, Li, and Bing]{videollama}
Hang Zhang, Xin Li, and Lidong Bing.
\newblock {Video-LLaMA: An Instruction-tuned Audio-Visual Language Model for Video Understanding}.
\newblock \emph{arXiv preprint arXiv:2306.02858}, 2023.

\bibitem[Zhang et~al.(2024)Zhang, Li, Liu, Lee, Gui, Fu, Feng, Liu, and Li]{llavanextvideo}
Yuanhan Zhang, Bo Li, haotian Liu, Yong~jae Lee, Liangke Gui, Di Fu, Jiashi Feng, Ziwei Liu, and Chunyuan Li.
\newblock {LLaVA-NeXT: A Strong Zero-shot Video Understanding Model}, 2024.

\bibitem[Zhao et~al.(2024)Zhao, Lu, Huo, Du, Yue, Guo, Wang, Chen, and Liu]{vnbench}
Zijia Zhao, Haoyu Lu, Yuqi Huo, Yifan Du, Tongtian Yue, Longteng Guo, Bingning Wang, Weipeng Chen, and Jing Liu.
\newblock {Needle In A Video Haystack: A Scalable Synthetic Framework for Benchmarking Video MLLMs}.
\newblock \emph{arXiv preprint}, 2024.

\bibitem[Zhao et~al.(2025)Zhao, Liu, Wu, Wang, Li, Wang, Teng, Liu, Cui, Wang, et~al.]{zhao2025clip}
Zihao Zhao, Yuxiao Liu, Han Wu, Mei Wang, Yonghao Li, Sheng Wang, Lin Teng, Disheng Liu, Zhiming Cui, Qian Wang, et~al.
\newblock Clip in medical imaging: A survey.
\newblock \emph{MIA}, 2025.

\bibitem[Zhu et~al.(2023)Zhu, Lin, Ning, Yan, Cui, Wang, Pang, Jiang, Zhang, Li, et~al.]{languagebind}
Bin Zhu, Bin Lin, Munan Ning, Yang Yan, Jiaxi Cui, HongFa Wang, Yatian Pang, Wenhao Jiang, Junwu Zhang, Zongwei Li, et~al.
\newblock {Languagebind: Extending Video-Language Pretraining to N-modality by Language-Based Semantic Alignment}.
\newblock \emph{arXiv preprint arXiv:2310.01852}, 2023.

\bibitem[Zou et~al.(2024)Zou, Yi, Li, and Li]{cls_closer}
Yixiong Zou, Shuai Yi, Yuhua Li, and Ruixuan Li.
\newblock A closer look at the cls token for cross-domain few-shot learning.
\newblock In \emph{NeurIPS}, 2024.

\end{thebibliography}
